\documentclass[%
        a4paper%
        ]{IEEEtran}
\usepackage{cite}
\usepackage{algorithmic}
\usepackage{graphicx}
\usepackage[caption=false,font=footnotesize]{subfig}
\usepackage{textcomp}
\usepackage[dvipsnames]{xcolor}
\usepackage{amsmath,amssymb,amsfonts}
\usepackage{mathtools}
\usepackage{tikz}
\usepackage{pgfplots}
\pgfplotsset{compat=newest}
\usepackage{booktabs}
\usepackage{lipsum}
\usepackage{multirow}
\usepackage{balance}
\usepackage{makecell}
\usepackage{tabularx}
\usepackage{bm}
\usepackage{hyperref}
\hypersetup{hidelinks=true}

\usepackage{siunitx}

\usetikzlibrary{spy}
\usetikzlibrary {arrows.meta,graphs,shapes.misc}
\usetikzlibrary{shapes.geometric}
\usetikzlibrary{positioning, calc, decorations.pathreplacing}
\usetikzlibrary{circuits.logic.US,tikzmark}
\usetikzlibrary{shapes}
\usetikzlibrary{shadows.blur}
\usetikzlibrary{positioning}
\usetikzlibrary{plotmarks}
\usetikzlibrary{shapes.geometric, arrows.meta, positioning, calc, fit, backgrounds}
\tikzset{>=latex}
\usepgfplotslibrary{colorbrewer}

\newcommand{\NNDET}{LLR-CNet}
\newcommand{\NNEST}{NBI-CNet}

\usepackage{soul}
\setul{0.5ex}{1.2pt}
\setstcolor{red}
\usetikzlibrary{external}
\hypersetup{
    colorlinks=true,
    linkcolor=blue,
    citecolor=green!70!black,
    filecolor=magenta,      
    urlcolor=cyan,
    }
    
\begin{document}

\title{Deep Learning for Joint Narrowband Interference Cancellation and Soft Demodulation in OFDM~Systems}

\author{
    \IEEEauthorblockN{Emmanouil Kavvousanos\IEEEauthorrefmark{1}, Francky Catthoor\IEEEauthorrefmark{2} and Vassilis Paliouras\IEEEauthorrefmark{1}}\\
    \IEEEauthorblockA{\IEEEauthorrefmark{1}Department of Electrical and Computer Engineering, University of Patras, Greece}\\
    \IEEEauthorblockA{\IEEEauthorrefmark{2}Department of Electrical and Computer Engineering, National Technical University of Athens, Greece}
}

\maketitle

\begin{abstract}
Narrowband interference (NBI) severely degrades orthogonal frequency-division multiplexing (OFDM) systems by corrupting subcarriers and rendering classical soft demodulation ineffective. Conventional compressed-sensing (CS) mitigation exhibits high sequential latency and leaves structured, non-Gaussian residuals that cause log-likelihood ratio (LLR) unreliability, decoder saturation, and severe error floors when employing classical Gaussian demappers. We resolve this pipeline mismatch using a unified deep learning framework for joint NBI cancellation and robust soft demodulation. First, NBI-CNet employs a physics-informed convolutional architecture to estimate NBI parameters and remove multi-tone interference in a single forward pass. Without requiring prior knowledge of the active interferer count, NBI-CNet reduces computational complexity by up to 60\% ($N{=}2048, Q{=}64$) compared to the state-of-the-art EOMP-IDS algorithm. Second, LLR-CNet acts as a structural whitener by mapping non-Gaussian post-mitigation residuals onto well-calibrated soft metrics. Simulations demonstrate that this joint framework eliminates the error floors inherent to traditional baselines across dense grids. Under severe interference ($\text{SIR}{=}{-}10$~dB), the pipeline operates within a $0.2$ to $0.5$~dB SNR margin of the optimal iterative baseline at a target block error rate (BLER) of $10^{-4}$. Under mild interference ($\text{SIR}{=}10$~dB) with heavy spectral overlap ($Q{=}12$), where classical greedy algorithms erroneously subtract valid data components and corrupt the payload, NBI-CNet avoids signal-peak confusion to deliver a coding gain exceeding $3$~dB. Finally, the architecture circumvents the $2{\times}10^{-4}$ error floor triggered by interferer-estimation errors, while its scale-invariant design enables robust generalization across arbitrary FFT sizes without retraining.
\end{abstract}

\section{Introduction}
\label{sec:intro}
\IEEEPARstart{T}{he} transition toward fifth-generation (5G), beyond-5G, and sixth-generation (6G) wireless networks features an exponential increase in device density and heterogeneous architectures designed for services like URLLC, mMTC, and eMBB \cite{qamarInterferenceManagementIssues2019, siddiquiURLLC5G6G2023, lohan5G6GNetworks2024}. This evolution integrates terrestrial cellular networks with UAVs \cite{shakhatrehSystematicReviewInterference2024}, satellite multibeam systems \cite{pengIntegratingTerrestrialSatellite2022}, and integrated sensing and communication \cite{niuInterferenceManagementIntegrated2025}. While these multi-tier layouts enhance global connectivity and spectrum utilization, they introduce severe interference challenges. As the RF spectrum becomes congested, the primary performance bottleneck shifts from a noise-limited to an interference-limited regime \cite{tushaInterferenceBurdenWireless2025}, necessitating robust detection and mitigation frameworks. This challenge is especially critical in the 6G FR3 upper mid-band (7.125--24.25~GHz), which offers substantial bandwidth but requires strict coexistence with incumbent satellite and defense services \cite{bazziUpperMidBandSpectrum2026, testolinaSpectrumSharingTerrestrial2025}. Dense terrestrial base stations can generate severe RFI toward satellite receivers via unintended sidelobe leakage and multipath reflections~\cite{testolinaSpectrumSharingTerrestrial2025}. Consequently, realizing the full potential of the FR3 spectrum demands multidimensional spectrum management to protect incumbents and mitigate inter-tier and intra-network interference \cite{bazziUpperMidBandSpectrum2026, cui6GWirelessCommunications2025}.

Within this congested landscape, narrowband interference (NBI) poses an acute threat to wideband communication systems \cite{giorgettiEffectNarrowbandInterference2005}, particularly those utilizing orthogonal frequency division multiplexing (OFDM) \cite{batraNarrowbandInterferenceMitigation2008, zhangMultipleInteractingNarrowband2020}. This threat natively arises where uncoordinated narrowband transmitters (such as dense IoT networks, legacy broadcasting, or industrial sensors) overlap with this wideband spectrum \cite{aygurNarrowbandInterferenceMitigation2025, husainDeepLearningBasedMultiTone2021}. This coexistence is increasingly prevalent in unlicensed bands, power line communications (PLC), and scenarios like NB-IoT sharing in-band LTE spectrum \cite{liuEliminatingNBIoTInterference2019, gaoJointImpulsiveNoise2025, martinezmarreroImprovingSoftDecoding2019}. In OFDM architectures, NBI disrupts subcarrier orthogonality. Because uncoordinated transmissions are typically asynchronous to the receiver's grid, their power inevitably leaks across adjacent subcarriers, degrading data recovery across a wide bandwidth \cite{batraNarrowbandInterferenceMitigation2008, kimNBISpectralLeakage2016}. This degradation is exacerbated when closely spaced tones interact, creating overlapping spectral aliasing that disrupts demodulation and limits classical suppression algorithms \cite{zhangMultipleInteractingNarrowband2020}. Navigating these environments requires advanced algorithms to identify and suppress interacting tones without relying on strict frequency-domain excision~\cite{huNarrowbandInterferenceCancellation2025, husainDeepLearningBasedMultiTone2021}.

Furthermore, while multi-tone NBI modeling often treats interference as a static, wide-sense stationary process within a single transmission frame \cite{aygurNarrowbandInterferenceMitigation2025}, this assumption may fail in practice. Physical-layer systems must account for the inherently time-varying statistics of the RF environment over extended periods \cite{gaoJointImpulsiveNoise2025, oyedareInterferenceSuppressionUsing2022}. Empirical measurements demonstrate that fundamental interference parameters (such as the number of active tones, center frequencies, fractional offsets, and instantaneous power levels) fluctuate dynamically from one OFDM symbol to the next due to changing channel conditions and uncoordinated source behavior \cite{huNarrowbandInterferenceCancellation2025}. Therefore, robust mitigation frameworks cannot rely exclusively on static parameter estimations; they must continuously adapt to this non-stationary behavior to effectively suppress dynamic interference and preserve underlying data \cite{gaoJointImpulsiveNoise2025, oyedareInterferenceSuppressionUsing2022}.

Current NBI mitigation strategies span time-domain processing, frequency-domain excision, sparse signal recovery \cite{kimNBISpectralLeakage2016, troppSignalRecoveryRandom2007, gaoJointImpulsiveNoise2025}, and machine learning \cite{aygurNarrowbandInterferenceMitigation2025, oyedareInterferenceSuppressionUsing2022, tushaInterferenceBurdenWireless2025}. Time-domain techniques often compromise between interference suppression and signal energy preservation \cite{batraNarrowbandInterferenceMitigation2008, liuEliminatingNBIoTInterference2019}. Conversely, frequency-domain excision nulls corrupted subcarriers based on energy thresholds, causing unavoidable data loss \cite{huNarrowbandInterferenceCancellation2025, aygurNarrowbandInterferenceMitigation2025}. To mitigate this, error correction strategies like log-likelihood ratio (LLR) erasure and soft-decision weighting enhance decoder resilience without modifying the transmission waveform \cite{husainDeepLearningBasedMultiTone2021, martinezmarreroImprovingSoftDecoding2019}.

To avoid data loss, compressive sensing (CS) exploits the frequency-domain sparsity of NBI. Algorithms like Orthogonal Matching Pursuit (OMP) reconstruct and subtract the interference vector without sacrificing spectral efficiency \cite{kimNBISpectralLeakage2016, liuEliminatingNBIoTInterference2019}. However, CS recovery algorithms require high computational complexity due to heavy matrix computations and often necessitate dedicated network resources, such as known frequency-domain elements or time-domain interference-free regions \cite{kimNBISpectralLeakage2016}. Furthermore, many CS-aided solutions possess a convergence-dependent rigid model structure that cannot cope well with dynamic wireless environments \cite{huNarrowbandInterferenceCancellation2025}.  Furthermore, in mobile environments with simultaneous impulsive noise and NBI, conventional CS constraints introduce prohibitive latency~\cite{gaoJointImpulsiveNoise2025}. Advanced variants such as OMP with Iterative Dichotomous Search (OMP-IDS) and Enhanced OMP-IDS (EOMP-IDS) achieve high interference cancellation depth \cite{huNarrowbandInterferenceCancellation2025}, serving as the primary comparative baselines for this work. However, beyond their high computational complexity and iterative nature, a major limitation of these methods is their strict dependency on a prior accurate estimation of the interferer count.

Recognizing these constraints, research has pivoted toward deep learning (DL) as a robust alternative capable of approximating complex, non-linear interference dynamics directly from raw data \cite{osheaIntroductionDeepLearning2017, heModelDrivenDeepLearning2019, oyedareInterferenceSuppressionUsing2022, lohan5G6GNetworks2024, liuEliminatingNBIoTInterference2019}. For interference mitigation, convolutional neural networks (CNNs) have been actively deployed to process baseband samples, allowing receivers to accurately locate narrowband interference in real-time with extremely low latency \cite{robinsonNarrowbandInterferenceDetection2023}. Furthermore, denoising autoencoders offer a powerful unsupervised approach to suppress heterogeneous interference by compressing the corrupted input into a latent-space representation and forcing the decoder network to filter out anomalous components \cite{oyedareInterferenceSuppressionUsing2022}. End-to-end DL models are also increasingly replacing conventional physical-layer detectors altogether; for instance, neural noncoherent receivers can now decode signals corrupted by both multi-user and narrowband interference without relying on explicit parameter estimations or manual thresholding \cite{sharmaDeepLearningNoncoherent2021}.

Beyond direct interference suppression, DL has fundamentally revolutionized downstream soft demodulation. Neural receivers can approximate optimal log-maximum a-posteriori (log-MAP) LLRs with a fraction of the computational complexity required by exact analytical implementations \cite{shentalMachineLLRningLearning2019, kavvousanosDesignImplementationLowComplexity2025}. This capability is particularly vital in correlated noise environments and under severe multi-tone interference, where 1D-CNNs successfully capture local correlation structures to produce highly accurate LLRs without explicitly computing the covariance matrix of the noise \cite{kavvousanosDesignImplementationLowComplexity2025}. Similarly, long short-term memory (LSTM) architectures have been utilized to classify affected subcarriers and dynamically weight their soft LLRs, significantly outperforming traditional energy detectors by preventing the severe information loss associated with hard erasure techniques \cite{husainDeepLearningBasedMultiTone2021}.

Despite these advancements, a critical gap remains: the interaction between imperfect NBI cancellation and downstream soft demodulation is fundamentally mismatched.
Both classical and DL cancellation algorithms inevitably leave behind structured, non-Gaussian residual interference. Classical soft demappers rely on strict additive white Gaussian noise assumptions, failing to capture the local correlation structures and high-amplitude bursts characteristic of non-ideal residuals \cite{shentalMachineLLRningLearning2019, kavvousanosDesignImplementationLowComplexity2025}. Consequently, processing this residual structure through a mismatched Gaussian demapper generates extreme LLR outliers \cite{kalyaniInterferenceMitigationTurboCoded2008}. Within modern forward error correction (FEC) schemes like low-density parity-check (LDPC) codes, these outliers severely disrupt decoding by acting as absorbing states that stall convergence. As demonstrated under multi-tone interference \cite{husainDeepLearningBasedMultiTone2021}, classical receivers feeding uncorrected LLRs directly into the decoder fail to resolve corrupted metrics, creating unavoidable error floors that compromise the entire FEC process.

To address all the limitations discussed above, this paper proposes a unified deep-learning framework that jointly performs NBI mitigation and robust soft demodulation in OFDM systems. The main contributions are:
\begin{enumerate}
    \item We propose \NNEST{}, a physics-informed convolutional network for parametric NBI estimation and cancellation. By modeling fractional spectral leakage without prior knowledge of the interferer count, it removes multi-tone interference with a near-constant, predictable FLOP count that falls well below the computational cost of EOMP-IDS in high-density environments.
    \item We introduce \NNDET{}, a lightweight neural LLR estimator tailored to non-Gaussian residual statistics. Leveraging both the suppressed signal and the estimated interference footprint, it acts as a structural whitener to produce well-calibrated soft information and eliminate severe LLR outliers.
    \item We demonstrate that the joint architecture eliminates the error floors inherent to conventional pipelines. It tracks optimal baselines closely under severe interference, while providing significant BLER gains under mild interference where greedy classical algorithms suffer from signal-peak confusion and erroneously destroy valid payload data.
    \item We show that the framework generalizes seamlessly across arbitrary FFT sizes and interference densities without retraining. Operating independently of the interferer count, it entirely circumvents the error floors triggered in classical compressed-sensing baselines by imperfect model-order estimation, supporting reliable, highly parallelized deployment.
\end{enumerate}

The remainder of this paper is organized as follows. Section~\ref{sec:system_model} presents the system model. Section~\ref{sec:classic_nbi} reviews classical compressed-sensing baselines. Section~\ref{sec:nbi_estimator} introduces \NNEST{}. Section~\ref{sec:residual_impact} details residual interference impacts, motivating the \NNDET{} estimator in Section~\ref{sec:llrnet}. Simulation and complexity analyses are in Sections~\ref{sec:simulation_results} and~\ref{sec:complexity}, followed by the conclusion in Section~\ref{sec:conclusion}.

\section{System and Interference Models}
\label{sec:system_model}
\subsection{System Model}

We consider a conventional cyclic-prefix orthogonal frequency-division multiplexing (CP-OFDM) system with $N$ subcarriers. Let $\mathbf{X} = [X_0, X_1, \dots, X_{N-1}]^T \in \mathbb{C}^{N \times 1}$ denote the vector of complex-valued data symbols mapped onto the subcarriers during one OFDM symbol interval.

The frequency-domain symbol vector $\mathbf{X}$ is transformed into the time domain via an inverse discrete Fourier transform (IDFT), yielding
\begin{equation}
s[n] = \frac{1}{\sqrt{N}} \sum_{k=0}^{N-1} X_k e^{j2\pi kn/N}, \quad n = 0, \dots, N-1.
\label{eq:ifft}
\end{equation}

To combat inter-symbol interference (ISI) caused by multipath propagation, a cyclic prefix (CP) of length $N_{\text{CP}}$ is appended to the beginning of each OFDM symbol. The transmitted time-domain signal becomes
\begin{equation}
s_{\text{CP}}[n] =
\begin{cases}
s[n {+} N {-} N_{\text{CP}}], & n = 0, \dots, N_{\text{CP}} {-} 1, \\
s[n {-} N_{\text{CP}}], & n = N_{\text{CP}}, \dots, N_{\text{CP}} {+} N {-} 1.
\end{cases}
\label{eq:cp_append}
\end{equation}

At the receiver, after CP removal, the received signal is sampled over $N$ samples and transformed back into the frequency domain using a discrete Fourier transform (DFT)
\begin{equation}
Y_k = \frac{1}{\sqrt{N}} \sum_{n=0}^{N-1} r[n] e^{-j2\pi kn/N}, \quad k = 0, \dots, N-1,
\label{eq:fft}
\end{equation}
where $r[n]$ denotes the received time-domain signal after CP removal.

Assuming that the CP length exceeds the maximum channel delay spread, the linear convolution between the transmitted signal and the channel impulse response is converted into a circular convolution. As a result, the frequency-domain received signal at the $k$th subcarrier can be written as
\begin{equation}
Y_k = H_k X_k + E_k + W_k,
\label{eq:freq_received}
\end{equation}
where $H_k$ is the channel frequency response, $E_k$ denotes the frequency-domain NBI component, and $W_k$ is the complex additive white Gaussian noise (AWGN) at subcarrier $k$.

Combining all subcarriers yields the vectorized model
\begin{equation}
\mathbf{Y} = \mathbf{H}\mathbf{X} + \mathbf{E} + \mathbf{W},
\label{eq:system_model}
\end{equation}
where $\mathbf{Y} \in \mathbb{C}^{N \times 1}$ is the received signal vector, $\mathbf{H} \in \mathbb{C}^{N \times N}$ is a diagonal matrix representing the frequency-domain channel response, $\mathbf{E} \in \mathbb{C}^{N \times 1}$ denotes the aggregate NBI component, and $\mathbf{W} {\sim}\kern3pt \mathcal{CN}(\mathbf{0}, \sigma_w^2\mathbf{I}_N)$ is the AWGN vector, where its entries are independent and identically distributed (i.i.d.) circularly symmetric complex Gaussian random variables with zero mean and variance $\sigma_w^2$. Here, $\mathbf{I}_N$ is the $N \times N$ identity matrix.

The objective of the receiver is to mitigate the impact of NBI prior to data detection, by estimation and cancellation of $\mathbf{E}$, and to subsequently produce well-calibrated soft information that is statistically matched to the residual disturbance, enabling reliable LDPC decoding.

\begin{figure*}[t]
\centering
\includegraphics[width=2\columnwidth]{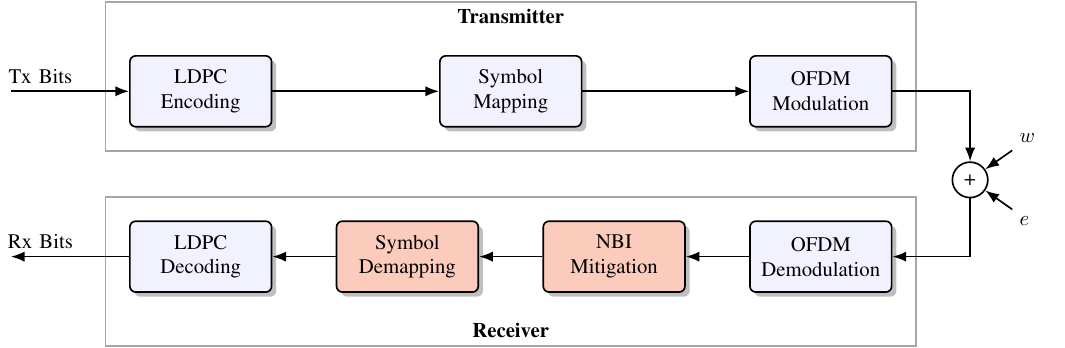}
\caption{Block diagram of the OFDM communication system with NBI mitigation.}
\label{fig:sys_model}
\end{figure*}

\subsection{Narrowband Interference Model}

We assume the presence of $Q$ independent narrowband interferers, each occupying a small fraction of the OFDM bandwidth. Typically, NBI is characterized by occupying no more than 5\% of the total available bandwidth \cite{huNarrowbandInterferenceCancellation2025, liuResearchKeyTechnologies2021}. The $q$th interferer is modeled in the time domain as a complex exponential
\begin{equation}
e_q[n] = g_q  e^{j\theta_q} \exp\left(j \frac{2\pi f_q n}{N}\right), \quad n = 0, \dots, N{-}1,
\label{eq:nbi_time}
\end{equation}
where $g_q$ is the interference amplitude (with power $P_q = g_q^2$), $f_q \in \mathbb{R}$ is the continuous-valued subcarrier frequency index, and $\theta_q$ is a random phase uniformly distributed over $[-\pi,\pi)$.

The subcarrier frequency index $f_q$ can be decomposed as
\begin{equation}
f_q = m_q + \alpha_q,
\end{equation}
where $m_q \in \{0,\dots,N-1\}$ is the integer subcarrier index and $\alpha_q \in [-\tfrac{1}{2}, \tfrac{1}{2})$ is the fractional frequency offset relative to the subcarrier spacing. When $\alpha_q = 0$, the interference is synchronous with the OFDM grid, whereas $\alpha_q \neq 0$ corresponds to asynchronous NBI.

Applying the DFT, the frequency-domain representation of the $q$th interferer at the $k$th subcarrier, $E_{q,k}$, can be derived in closed form using the sum of a geometric progression
\begin{equation}
\begin{split}
E_{q,k} &= \frac{1}{\sqrt{N}} \sum_{n=0}^{N-1} e_q[n] e^{-j2\pi kn/N} \\
&= \frac{g_q}{\sqrt{N}} e^{j\theta_q} e^{j\pi(f_q - k)\frac{N-1}{N}} \underbrace{\frac{\sin(\pi(f_q - k))}{\sin(\frac{\pi}{N}(f_q - k))}}_{\mathcal{D}_N(f_q - k)}.
\end{split}
\label{eq:nbi_freq_closed_form}
\end{equation}

In \eqref{eq:nbi_freq_closed_form}, the ratio of the sine functions represents the periodic sinc function, fundamentally known as the Dirichlet kernel, denoted as $\mathcal{D}_N(f_q {-} k)$. This kernel mathematically governs the spectral envelope and leakage behavior of the interference across the subcarrier grid.

If the NBI is strictly synchronous ($\alpha_q = 0$), the Dirichlet kernel evaluates to zero for all subcarriers except at $k = m_q$, where it limits to a peak amplitude of $N$. This concentrates all interference energy exclusively onto the single subcarrier $m_q$, behaving exactly like a Kronecker delta. However, in the asynchronous case ($\alpha_q \neq 0$), the continuous nature of the Dirichlet kernel causes spectral leakage; the interference energy spreads across multiple adjacent subcarriers, creating characteristic high-power spectral tails around the central fractional frequency.

The aggregate NBI vector affecting the received signal is the linear superposition of all $Q$ interferers
\begin{equation}
\mathbf{E} = \sum_{q=1}^{Q} \mathbf{E}_q,
\label{eq:total_nbi}
\end{equation}
where $\mathbf{E}_q = [E_{q,0}, \dots, E_{q,N-1}]^T$. Due to the highly localized main-lobe and rapidly decaying side-lobes of the underlying Dirichlet kernel, the aggregate interference vector $\mathbf{E}$ inherently exhibits a sparse or block-sparse structure in the frequency domain. %

\subsection{Problem Formulation and Assumptions}

In this work, we isolate the performance of the NBI parameter estimation, cancellation, and soft demodulation modules. Evaluating these components independently clarifies their standalone performance limits without the influence of errors from subsequent receiver stages. We assume an AWGN channel for the signal of interest ($H_k {=} 1, \forall k$), which mirrors the effective background noise floor of a post-equalization signal in practical deployments. Fading or multipath propagation effects experienced by the NBI sources are inherently absorbed into their received amplitude $g_q$ and phase $\theta_q$ parameters, preserving the mathematical generality of the interference model.

Regarding the interference profile, we assume a continuous yet dynamically time-varying NBI environment where the active interferer count $Q$ and its constituent parameters vary independently from one OFDM symbol to the next. Rather than a limiting constraint, this symbol-to-symbol variance represents a worst-case scenario designed to stress-test the receiver. While stationary or slow-fading NBI allows conventional demappers to track and mitigate residual statistics over wider windows (e.g., a 14-symbol slot), restricting parameter persistence to a single isolated symbol prevents multi-symbol statistical averaging. This setup intentionally exposes the tracking limitations of classical architectures and evaluates the proposed framework under maximum temporal volatility~\cite{kalyaniInterferenceMitigationTurboCoded2008}.

\section{Classic NBI Cancellation via Compressed Sensing}
\label{sec:classic_nbi}
To establish a rigorous mathematical foundation and provide baseline techniques for comparison with our proposed neural architectures, this section briefly reviews state-of-the-art NBI mitigation techniques based on CS. In strict accordance with our system assumptions, these CS algorithms operate independently on isolated OFDM symbols, recovering the interference profile without relying on temporal tracking. To mitigate NBI without sacrificing spectral efficiency (unlike simple frequency-domain excision which completely nulls useful data subcarriers), CS algorithms leverage the inherent structural sparsity of the interference \cite{troppSignalRecoveryRandom2007,kimNBISpectralLeakage2016,huNarrowbandInterferenceCancellation2025}. In the frequency domain, NBI manifests as a few high-energy peaks, meaning the interference vector $\mathbf{E} \in \mathbb{C}^{N \times 1}$ can be represented as a sparse linear combination of atoms from a redundant dictionary. 

\subsection{Sparse Representation and Dictionary Construction}
The frequency-domain NBI vector $\mathbf{E} \in \mathbb{C}^{N \times 1}$ can be modeled as a sparse linear combination of atoms from a redundant sensing matrix, expressed as
\begin{align}
    \mathbf{E} = \bm{\Phi} \mathbf{d},
    \label{eq:e_phi_d}
\end{align} 
where $\bm{\Phi} \in \mathbb{C}^{N \times M}$ is the dictionary and $\mathbf{d} \in \mathbb{C}^{M \times 1}$ is the sparse coefficient vector~\cite{kimNBISpectralLeakage2016}. To directly link this compressed-sensing framework to the physical multi-tone NBI parameters defined in \eqref{eq:nbi_freq_closed_form}, the dictionary $\bm{\Phi}$ is constructed by concatenating candidate narrowband spectral signatures across a dense frequency grid
\begin{equation}
\bm{\Phi} = [\bm{\phi}_1, \bm{\phi}_2, \dots, \bm{\phi}_M].
\end{equation}
Each atom $\bm{\phi}_i$ represents the exact frequency-domain leakage profile of a hypothetical single-tone interferer at subcarrier frequency index $f_i$. To precisely mirror the physical signal model, the $k$th subcarrier element of atom $\bm{\phi}_i$ is defined by the Dirichlet kernel signature
\begin{equation}
[\bm{\phi}_i]_k = \frac{1}{\sqrt{N}} e^{j\pi(f_i - k)\frac{N-1}{N}} \mathcal{D}_N(f_i - k).
\end{equation}

To account for asynchronous NBI where the continuous interferer frequency $f_q$ does not align with the standard DFT grid, the dictionary is oversampled such that $M \gg N$ \cite{kimNBISpectralLeakage2016}. This grid redundancy allows the receiver to minimize grid-mismatch errors, ensuring that the continuous spectral leakage of the true interference signal is closely approximated by the closest discrete atoms in $\bm{\Phi}$.

The corresponding sparse coefficient vector $\mathbf{d} {=} [d_1, d_2, \dots, d_M]^T$ encapsulates the complex power of the active interferers. If a physical interferer $q$ aligns with the dictionary grid frequency $f_i$, its respective coefficient becomes non-zero and maps to $d_i = g_q e^{j\theta_q}$, thereby absorbing the physical amplitude and phase of the source. The estimated NBI vector $\hat{\mathbf{E}}$ can then be reconstructed as a linear combination of the active atoms
\begin{align}
    \hat{\mathbf{E}} = \sum_{i=1}^{M} d_i \bm{\phi}_i.
    \label{eq:cs-recon}
\end{align}
Finally, interference mitigation is performed by subtracting this reconstructed profile from the received frequency-domain signal vector $\mathbf{Y}$, yielding the cleaned signal
\begin{align}
    \tilde{\mathbf{Y}} = \mathbf{Y} - \hat{\mathbf{E}}.
\end{align}

\subsection{The Orthogonal Matching Pursuit Algorithm and Extensions}

Given the received signal $\mathbf{Y}$ and an estimate $\widehat{Q}$ of the interferer count, the classical Orthogonal Matching Pursuit (OMP) algorithm \cite{troppSignalRecoveryRandom2007, kimNBISpectralLeakage2016} recovers the sparse vector $\mathbf{d}$ through a greedy iterative process. At each iteration $q$, the algorithm identifies the dictionary atom $\boldsymbol{\phi}_i$ exhibiting the maximum absolute correlation with the current residual $\mathbf{R}^{(q-1)}$
\begin{equation}
\label{eq:omp_select}
i^{(q)} = \arg\max_{i \in \{1, \dots, M\}} \left| \langle \mathbf{R}^{(q-1)}, \boldsymbol{\phi}_i \rangle \right|.
\end{equation}
The selected index is appended to the active support set $\mathcal{S}^{(q)} = \{i^{(1)}, \dots, i^{(q)}\}$. To guarantee orthogonality, the coefficients $\hat{\mathbf{d}}_{\mathcal{S}^{(q)}}$ are updated via least-squares projection onto the subspace spanned by the current support
\begin{equation}
\label{eq:omp_invert}
\hat{\mathbf{d}}_{\mathcal{S}^{(q)}} = (\boldsymbol{\Phi}_{\mathcal{S}^{(q)}}^H \boldsymbol{\Phi}_{\mathcal{S}^{(q)}})^{-1} \boldsymbol{\Phi}_{\mathcal{S}^{(q)}}^H \mathbf{Y}.
\end{equation}
The residual is then updated as
\begin{align}
    \mathbf{R}^{(q)} = \mathbf{Y} - \boldsymbol{\Phi}_{\mathcal{S}^{(q)}} \hat{\mathbf{d}}_{\mathcal{S}^{(q)}},
\end{align}
and the process repeats until $\widehat{Q}$ components are isolated.

To mitigate the discrete grid mismatch problem inherent to standard OMP, the Iterative Dichotomous Search (IDS) extension \cite{huNarrowbandInterferenceCancellation2025} refines the coarse frequency estimate $\hat{f}_{\text{coarse}}$. Treating the continuous correlation function $\rho(f) = |\langle \mathbf{R}, \boldsymbol{\phi}(f) \rangle|$ as unimodal, IDS iteratively halves a localized search interval over a fixed number of steps. A semi-closed-form interpolation based on the final correlation values then calculates the continuous-valued frequency without fine-grid computational overhead. Furthermore, in multi-tone scenarios where adjacent spectral leakage peaks overlap, the EOMP framework \cite{huNarrowbandInterferenceCancellation2025} introduces an iterative decoupling stage. EOMP isolates each interferer $q \in \{1, \dots, \widehat{Q}\}$ by subtracting all other active estimates from the received observation
\begin{equation}
\label{eq:eomp_decouple}
\mathbf{R}_{\text{ref}, q} = \mathbf{Y} - \sum_{j \neq q} \boldsymbol{\phi}_{i^{(j)}} \hat{d}_{i^{(j)}}.
\end{equation}
IDS refinement is subsequently applied to this decoupled residual $\mathbf{R}_{\text{ref}, q}$ to eliminate multi-interferer bias, repeating the subtraction and refinement steps until convergence.

Despite improving cancellation depth, these classical CS frameworks introduce notable algorithmic constraints that impact processing efficiency. First, the greedy search in \eqref{eq:omp_select} must repeat sequentially across $Q$ discrete steps, blocking parallel hardware execution. Second, the subspace projection in \eqref{eq:omp_invert} demands a high-precision matrix pseudo-inversion $(\boldsymbol{\Phi}_{\mathcal{S}^{(q)}}^H \boldsymbol{\Phi}_{\mathcal{S}^{(q)}})^{-1}$ that expands in dimensionality at every single iteration $q$, generating a heavy, non-linear processing overhead. This latency expands further under EOMP-IDS because the subtraction step in \eqref{eq:eomp_decouple} embeds these operations inside an outer refinement loop, forcing repetitive on-the-fly atom generation that scales poorly with both the interferer count $Q$ and the OFDM grid size $N$. Finally, because loop execution relies entirely on the accuracy of the prior model order estimate $\widehat{Q}$, any estimation error destabilizes the pipeline: underestimating $\widehat{Q}$ terminates processing prematurely and leaves high-power interferers unmitigated, while overestimating forces \eqref{eq:omp_invert} and \eqref{eq:eomp_decouple} to project background noise onto the active support, actively destroying valid data subcarriers. These predictable latency and model-dependency limitations motivate the single-shot, parallelized neural estimation framework proposed in the next section.

\section{Neural Network-based NBI Cancellation}
\label{sec:nbi_estimator}

To overcome the limitations of the classical iterative algorithms discussed in Section~\ref{sec:classic_nbi}, we propose a parallelized, physics-informed CNN framework. While this DL approach is conceptually inspired by the physical reconstruction logic of traditional sparse recovery, its underlying architecture constitutes a fundamental mathematical reformulation that departs from conventional CS theory. Rather than relying on iterative grid-searches and sequential, on-the-fly atom refinement, our framework utilizes a deep neural network to predict continuous parametric states, dynamically synthesizing the localized interference profile. Consequently, beyond the high-level functional and structural connections explicitly detailed in this section, direct theoretical links to classical CS properties cannot be extended to this DL formulation.

Furthermore, unlike traditional DL approaches that operate as black-box models attempting to map the corrupted observation directly to a clean signal without explicitly considering the underlying signal physics~\cite{oyedareInterferenceSuppressionUsing2022, rockComplexSignalDenoising2019}, our architecture is designed as a parametric convolutional estimator coupled with a differentiable analytical reconstruction layer. By evaluating the entire spectrum in a single forward pass, the network explicitly extracts the continuous physical properties of the multitone NBI (gain, fractional frequency offset, and phase) for all subcarriers simultaneously. These predicted parameters are then fed directly into the reconstruction layer to analytically synthesize the exact interference waveform.

\subsection{NBI-CNet Architecture}
In order to address the spectral leakage caused by NBI in OFDM systems, we introduce \NNEST{}, a deep CNN architecture. Operating as a non-linear parameter estimator, \NNEST{} is designed to deduce the physical characteristics of the complex interference vector $\mathbf{E} \in \mathbb{C}^{N \times 1}$ directly from the frequency-domain observation $\mathbf{Y}$ established in \eqref{eq:system_model}.

The network accepts two inputs: the noisy, NBI-corrupted frequency-domain observation $\mathbf{Y} \in \mathbb{C}^{N \times 1}$, and the scalar thermal noise variance $\sigma_w^2$. To ensure scale-invariance across arbitrary FFT sizes ($N$) and maintain gradient stability, the complex input signal is normalized to a power-invariant representation
\begin{equation}
    \mathbf{Y}_\text{in} = \frac{\mathbf{Y}}{\sqrt{N}}.
\end{equation}

To process the complex signal within a real-valued neural network architecture, $\mathbf{Y}_\text{in}$ is decoupled into its real and imaginary components. The initial spatial feature map $\bm{\chi}^{(0)} \in \mathbb{R}^{N \times 2}$, which is fed into the convolutional feature extractor, is formed by concatenating these components, as
\begin{equation}
    \bm{\chi}^{(0)} = \left[ \Re\{\mathbf{Y}_{\text{in}}\}, \Im\{\mathbf{Y}_{\text{in}}\} \right].
\end{equation}
Meanwhile, the raw scalar noise variance $\sigma_w^2$, which is physically independent of the FFT size, bypasses the feature extraction stage and is reserved for concatenation in the subsequent state-conditioning dense layers.

The architecture of \NNEST{} consists of three main stages (see Fig.~\ref{fig:mt-cnn}):
\begin{enumerate}
    \item \textbf{Feature Extraction:} To capture the spectral leakage tails, we employ a series of $L$ 1D convolutional (Conv1D) layers. Let $\bm{\chi}^{(l)} \in \mathbb{R}^{N \times F_l}$ denote the intermediate feature map at layer $l \in \{1, \dots, L\}$, where $F_l$ is the number of filters. To mitigate spectral wrap-around effects at the band edges, circular padding is applied to the input feature map $\bm{\chi}^{(l-1)}$ prior to each convolution (see Section~\ref{subsec:circ-pad}), such that
    \begin{equation}
        \bm{\chi}^{(l)} = \text{ReLU}\left( \text{Conv1D}\left( \text{CircPad}\left(\bm{\chi}^{(l-1)}\right) \right) \right).
    \end{equation}
    
    \item \textbf{Noise Conditioning:} The final extracted feature map $\bm{\chi}^{(L)}$ is concatenated with the noise variance $\sigma^2_w$ across the channel dimension, forming the conditioned state tensor $\bm{\chi}_{\text{cond}} \in \mathbb{R}^{N \times (F_L + 1)}$. This conditions the dense layers on the current signal-to-noise ratio (SNR), allowing the network to dynamically adjust its sparsity and detection thresholds.
    
    \item \textbf{Physics-Constrained Multi-Head Output:} The conditioned state $\bm{\chi}_{\text{cond}}$ is split into three independent branches of pointwise dense layers, each dedicated to predicting a specific physical parameter of the NBI model for all $N$ subcarriers simultaneously. To ensure the network outputs remain physically valid, specialized activation functions are applied to the final layer of each head:
    \begin{enumerate}
        \item \textbf{Gain ($\hat{\mathbf{g}}$):} Yields the predicted amplitude vector $\hat{\mathbf{g}} {=} [\hat{g}_0, \dots, \hat{g}_{N-1}]^T \in \mathbb{R}^N$ using a ReLU activation to enforce non-negativity.
        \item \textbf{Frequency Offset ($\hat{\boldsymbol{\alpha}}$):} Yields the predicted fractional sub-bin offset vector $\hat{\boldsymbol{\alpha}} {=} [\hat{\alpha}_0, \dots, \hat{\alpha}_{N-1}]^T \in \mathbb{R}^N$ using a scaled hyperbolic tangent activation, $0.5 \tanh(\cdot)$, strictly bounding the offsets to the physical limits of $[-0.5, 0.5]$ subcarrier spacings.
        \item \textbf{Phase ($\hat{\boldsymbol{\theta}}$):} To avoid gradient discontinuities caused by phase wrap-around at $\pm \pi$, the network outputs a 2D Cartesian tensor $\bm{\chi}_{\theta} \in \mathbb{R}^{N \times 2}$. This tensor is strictly $L_2$-normalized across its channels to map features onto a unit circle, implicitly providing the phase parameters via $[\cos(\hat{\theta}_k), \sin(\hat{\theta}_k)]^{T}$ for each subcarrier $k$.
    \end{enumerate}
\end{enumerate}

\subsection{Interference Reconstruction}

To bridge the gap between the predicted parameters and final mitigation stage, \NNEST{} implements a parameter-free, fully differentiable reconstruction layer. Leveraging the mathematical duality of the signal model, this layer enables equivalent formulations in either the time or frequency domain.

For each subcarrier $k \in \{0, \dots, N-1\}$, the network's predicted fractional frequency offset $\hat{\alpha}_k$ is used to define the continuous interference frequency as $\hat{f}_k = k + \hat{\alpha}_k$. Because the amplitude head uses a ReLU activation, unoccupied subcarriers inherently evaluate to zero, natively enforcing structural sparsity without an explicit active-set threshold.

Following \eqref{eq:nbi_time}, the estimated time-domain NBI sequence $\hat{e}[n]$ is synthesized as a linear superposition over all subcarriers
\begin{equation}
    \hat{e}[n] = \sum_{k=0}^{N-1} \hat{g}_k e^{j\hat{\theta}_k} \exp\kern-2pt
    \left(\kern-2pt j \frac{2\pi\kern-1pt \hat{f}_k n}{N}\kern-2pt
    \right), \quad n = 0, \dots, N{-}1.
    \label{eq:time_recon}
\end{equation}

Alternatively, the frequency-domain NBI vector $\hat{\mathbf{E}}$ can be reconstructed directly in closed form. Leveraging the linear superposition principle from \eqref{eq:total_nbi}, the aggregate interference evaluated at any specific subcarrier bin $m$ is synthesized analytically via \eqref{eq:nbi_freq_closed_form} as
\begin{equation}
    \hat{E}_m = \sum_{k=0}^{N-1} \frac{\hat{g}_k}{\sqrt{N}} e^{j\hat{\theta}_k} e^{j\pi(\hat{f}_k - m)\frac{N-1}{N}} \mathcal{D}_N(\hat{f}_k - m).
    \label{eq:freq_recon}
\end{equation}

To explicitly demonstrate the functional equivalence between this neural reconstruction and the classical CS formulation from \eqref{eq:e_phi_d} in Section~\ref{sec:classic_nbi}, \eqref{eq:freq_recon} can be recast as a dynamic matrix-vector product. Let us define a complex sparse coefficient vector $\hat{\mathbf{d}} \in \mathbb{C}^{N \times 1}$, where the $k$th element encapsulates the predicted amplitude and phase of the corresponding subcarrier head, such that $\hat{d}_k = \hat{g}_k e^{j\hat{\theta}_k}$. 

Simultaneously, the predicted fractional offsets $\hat{\boldsymbol{\alpha}}$ are used to synthesize a dynamic, parameter-dependent dictionary matrix $\hat{\boldsymbol{\Phi}} \in \mathbb{C}^{N \times N}$, defined as $\hat{\boldsymbol{\Phi}} = [\hat{\boldsymbol{\phi}}_0, \dots, \hat{\boldsymbol{\phi}}_{N-1}]$. The $m$th frequency-domain bin of the $k$th continuous dictionary atom is evaluated analytically at $\hat{f}_k = k + \hat{\alpha}_k$ using the Dirichlet kernel as
\begin{equation}
    [\hat{\boldsymbol{\phi}}_k]_m = \frac{1}{\sqrt{N}} e^{j\pi(\hat{f}_k - m)\frac{N-1}{N}} \mathcal{D}_N(\hat{f}_k - m).
\end{equation}

The aggregate frequency-domain interference is then reconstructed via the linear combination
\begin{equation}
    \hat{\mathbf{E}} = \hat{\boldsymbol{\Phi}} \hat{\mathbf{d}}.
    \label{eq:freq_recon_matrix}
\end{equation}

\begin{figure}[tb]
    \centering
\includegraphics[width=1\columnwidth]{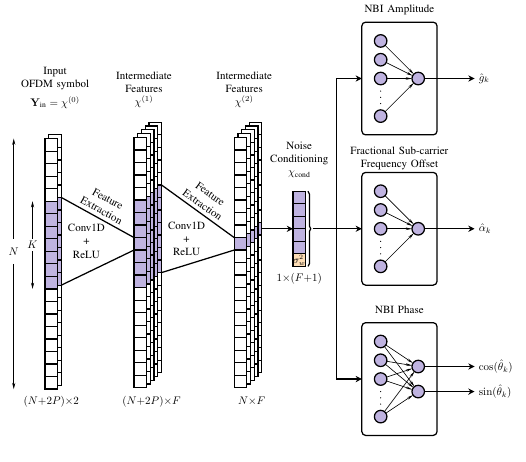}
    \caption{Model Architecture of \NNEST{}.}
    \label{fig:mt-cnn}
\end{figure}

By dynamically generating a $N {\times} N$ dictionary mapped directly to the continuous frequency domain, \NNEST{} mathematically mirrors the physical synthesis of the CS methods discussed in Section~\ref{sec:classic_nbi}. Finally, the estimated NBI vector is directly subtracted from the received signal, executing the final mitigation step as $\tilde{\mathbf{Y}} = \mathbf{Y} - \hat{\mathbf{E}}$. Functionally, this single-shot cancellation represents the highly parallelized equivalent of the terminal classical residual $\mathbf{R}^{(\widehat{Q})}$ of OMP-IDS. The \NNEST{} architecture is detailed in Fig.~\ref{fig:mt-cnn} and Table~\ref{tab:nnest_arch}.

\begin{table}[tb]
    \caption{\NNEST{} Architecture and Parameter Count}
    \label{tab:nnest_arch}
    \centering
    \includegraphics[width=1\columnwidth]{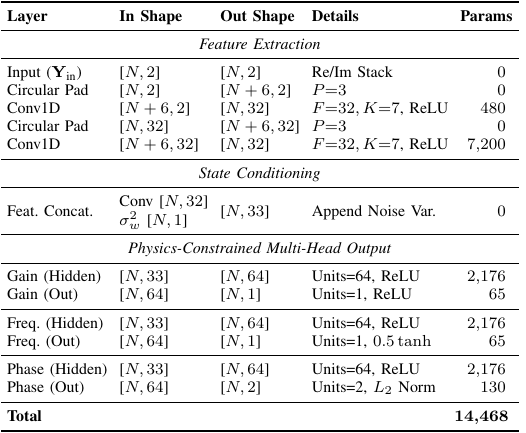}
\end{table}

\subsection{Circular Padding and Spectral Wrap-Around}
\label{subsec:circ-pad}
Because asynchronous NBI frequencies are continuous ($\alpha {\neq} 0$), their power leaks across adjacent subcarriers. When the interference frequency $m+\alpha$ is located near the DC or band edges, the interference naturally wraps around the discrete spectrum boundaries \cite{kimNBISpectralLeakage2016}. 

Standard zero-padding in convolutional layers destroys this physical continuity, introducing artificial boundary artifacts. To preserve the inherent circularity of the OFDM frequency domain, each 1D convolutional layer $l$ applies a circular padding operation prior to the convolution.

Let $K_l$ denote the kernel size of the $l$th convolutional layer, requiring a padding size of $P = (K_l - 1)/2$ on both spatial boundaries. Given an input sequence $\bm{\chi}^{(l)} \in \mathbb{R}^{N \times F_l}$ with $F_l$ feature channels, the circularly padded sequence $\bm{\chi}_{\text{pad}}^{(l)} \in \mathbb{R}^{(N+2P) \times F_l}$ is constructed using modulo indexing. Specifically, the element at spatial index $m$ is defined as
\begin{equation}
    \left[ \bm{\chi}_{\text{pad}}^{(l)} \right]_{\mathrlap{m}} = \bm{\chi}^{(l)}_{(m - P) \bmod N}, \quad m = 0, 1, \dots, N{+}2P{-}1.
\end{equation}

\begin{figure}[tb]
\includegraphics[width=1\columnwidth]{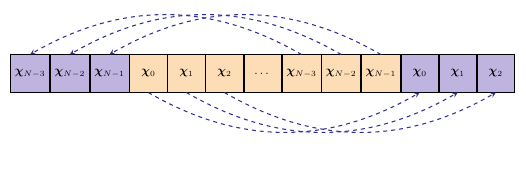}
\caption{Circular Padding for improved spectral leakage identification.}
\label{fig:padding}
\end{figure}

This operation effectively prepends the last $P$ subcarriers to the beginning of the sequence and appends the first $P$ subcarriers to the end, ensuring that convolutional filters spanning the band edges correctly observe the wrapped spectral leakage. An illustration of this circular padding mechanism is provided in Fig.~\ref{fig:padding}.

\subsection{Hybrid Loss Formulation}
The network is trained using a hybrid supervised loss function that balances direct parameter mapping with physical reconstruction accuracy. Because multitone NBI is highly sparse, the loss function is heavily masked to prevent gradient collapse from the vast majority of interference-free subcarriers.

Let $\mathcal{K} = \{0, 1, \dots, N-1\}$ denote the set of all OFDM subcarrier indices. We partition $\mathcal{K}$ into two mutually exclusive subsets: the active set $\mathcal{M}_{1}$, containing indices occupied by NBI, and the inactive set $\mathcal{M}_{0}$, containing interference-free indices. Formally, these are defined based on the ground-truth NBI amplitude $g_{k}$ as
\begin{align}
    \mathcal{M}_{1} &= \{ k \in \mathcal{K} \mid g_{k} > 0 \} \\
    \mathcal{M}_{0} &= \{ k \in \mathcal{K} \mid g_{k} = 0 \}.
\end{align}
The total loss $\mathcal{L}_{\text{total}}$ is a superposition of four components designed to treat these subsets appropriately:

\begin{enumerate}
    \item \textbf{Global Gain Loss}: Evaluated over the entire spectrum to teach the network peak localization, given by the Mean Squared Error (MSE)
    \begin{align*}
        \mathcal{L}_{g} = \frac{1}{N} \sum_{k=0}^{N-1} (g_{k} - \hat{g}_k)^2.
    \end{align*}
    
    \item \textbf{Masked Parameter Loss}: Frequency offset and phase are only physically meaningful where NBI exists. Therefore, their MSE are computed strictly over the active subset~$\mathcal{M}_{1}$
    \begin{align*}
    \mathcal{L}_{\alpha} &= \frac{1}{|\mathcal{M}_{1}|} \sum_{k \in \mathcal{M}_{1}} (\alpha_{k} - \hat{\alpha}_k)^2,\\
    \mathcal{L}_{\theta} &= \frac{1}{|\mathcal{M}_{1}|} \sum_{k \in \mathcal{M}_{1}} \big|e^{j\theta_{k}} - e^{j\hat{\theta}_k}\big|^2,
    \end{align*}
    where $|\mathcal{M}|$ denotes the cardinality of set $\mathcal{M}$.

    \item \textbf{Structural Sparsity Penalty}: Because true multitone NBI is structurally sparse in the frequency domain, the network must learn to output strict zero-gain values for unoccupied subcarriers. To enforce this and suppress noise-induced false positives, an $L_1$ penalty is applied solely to the inactive subset $\mathcal{M}_{0}$
    \begin{equation*}
                \mathcal{L}_{\text{sparse}} = \frac{1}{|\mathcal{M}_{0}|} \sum_{k \in \mathcal{M}_{0}} |\hat{g}_k|.
    \end{equation*}

\item \textbf{Physical Reconstruction Loss}: While the parameter-specific losses evaluate the network's outputs on an isolated, per-subcarrier basis, they do not capture the highly coupled physical relationship between amplitude, phase, and frequency offset. To enforce consistency, this metric evaluates the MSE between the true time-domain interference sequence $e[n]$ and the network's differentiable analytical reconstruction $\hat{e}[n]$ defined in~\eqref{eq:time_recon},
\begin{equation*}
    \mathcal{L}_{\text{recon}} = \frac{1}{N} \sum_{n=0}^{N-1} \left| e[n] - \hat{e}[n] \right|^2.
\end{equation*}
By penalizing the synthesized time-domain waveform directly, this term bridges any gaps left by the independent parameter losses and provides gradients that jointly optimize all network heads.
\end{enumerate}

The final objective optimized via gradient descent is %
\begin{align}
    \mathcal{L}_{\text{total}} = \mathcal{L}_{g} + \mathcal{L}_{\alpha} + \mathcal{L}_{\theta} + \lambda \mathcal{L}_{\text{sparse}} + \mathcal{L}_{\text{recon}},
    \label{eq:propobj}
\end{align}
where $\lambda$ acts as a sparsity regularization factor. The proposed objective~\eqref{eq:propobj} ensures that the network prioritizes sub-bin accuracy on the active tones while strictly suppressing false-positive NBI estimates across interference-free subcarriers.

\subsection{Training Procedure and Implementation Details}
\label{subsec:impl-details}
The proposed NBI estimator and the simulated communication system were implemented in TensorFlow \cite{abadi2016tensorflowlargescalemachinelearning} using the Sionna library \cite{sionna} for GPU-accelerated physical-layer operations. The computational environment for all training and evaluation routines was powered by an NVIDIA DGX Spark integrating the GB10 Grace Blackwell Superchip \cite{skende171NVIDIAGB102026}.%

To prevent overfitting and guarantee robust generalization across a continuous parameter space, the training dataset is generated dynamically on-the-fly. Rather than relying on a static, pre-computed dataset, a new mini-batch of OFDM symbols is procedurally synthesized at every training step. This continuous generation of data ensures that the network encounters unique noise and interference realizations during every iteration.

The simulated system employs 16-QAM modulation with a $N {=} 256$ subcarrier grid over an AWGN channel. To condition the network across a highly diverse range of spectral environments, the physical parameters of each batch are drawn from uniform distributions. The ambient SNR varies between $7$ and~$15$ dB, while the signal-to-interference ratio (SIR) spans an extreme dynamic range from $-30$ to $10$ dB. The number of active multi-tone NBI sources per symbol is randomized as $Q \sim \mathcal{U}(0, 8)$. 

To maintain the structural sparsity of the multitone interference model, a minimum NBI distance constraint of $d_m{=}2$ subcarriers is strictly enforced during generation. This distance dictates that the integer center frequencies ($m_q$) of any two active interferers must be separated by at least two discrete subcarrier indices. This constraint prevents independent narrowband sources from completely overlapping and collapsing into an indistinguishable, continuous wideband cluster. Furthermore, to accurately model asynchronous interference and induce realistic spectral leakage across this grid, the fractional frequency offsets $\alpha$ are drawn independently from $\mathcal{U}(-0.5, 0.5)$.

\NNEST{} was trained for $120,000$ steps utilizing the Adam optimizer with a learning rate of $\eta = 10^{-3}$ and a batch size of $256$. The sparsity regularization weight in the hybrid loss function was determined empirically and set to $\lambda = 0.3$. The complete set of simulation and training hyperparameters is summarized in Table~\ref{tab:training_params}.

\begin{table}[tb]
\centering
\caption{NBI Estimator Training Parameters}
\label{tab:training_params}
\includegraphics[width=1\columnwidth]{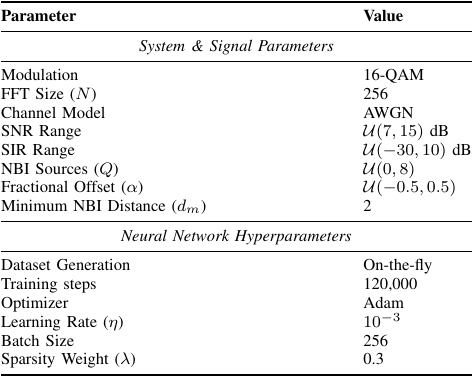}
\end{table}

\section{Impact of Residual Interference on LLR Estimation and LDPC Decoding}
\label{sec:residual_impact}

In conventional OFDM receivers, soft demodulation is typically performed under the assumption that the post-equalization disturbance consists solely of AWGN. Under this assumption, the LLRs for the coded bits are computed based on a Gaussian likelihood model with a known or estimated variance $\sigma_w^2$. However, when NBI is only partially mitigated, the residual interference $\tilde{\mathbf{E}} = \mathbf{E} - \hat{\mathbf{E}}$ introduces non-Gaussian distortion that violates this assumption and degrades LLR reliability.

Let $\tilde{Y}_k$ denote the interference-suppressed signal on the $k$th subcarrier after cancellation
\begin{equation}
\tilde{Y}_k = H_k X_k + \tilde{E}_k + W_k,
\end{equation}
where $\tilde{E}_k$ represents the residual NBI component and $W_k \sim \mathcal{CN}(0,\sigma_w^2)$ denotes AWGN. In the presence of residual interference, the effective disturbance $Z_k = \tilde{E}_k + W_k$ is generally neither Gaussian nor independent across subcarriers.

Despite this, conventional LLR computation assumes
\begin{equation}
p(\tilde{Y}_k | X_k) \approx \frac{1}{\pi \sigma_w^2} 
\exp\!\left(-\frac{|\tilde{Y}_k - H_k X_k|^2}{\sigma_w^2}\right),
\end{equation}
which ignores the residual interference term $\tilde{E}_k$. As a result, the resulting LLRs are mismatched with respect to the true channel statistics.

For a bit $b_{k,m}$ associated with symbol $X_k$, the classical max-log LLR is computed as
\begin{equation}
L(b_{k,m}) \approx
\frac{1}{\sigma_w^2}\kern-2pt
\bigg(
\min_{X \in \mathcal{X}_m^0}\!|\tilde{Y}_k {-} H_k X|^{\mathrlap{2}}
{-}
\min_{\mathclap{X \in \mathcal{X}_m^1}}|\tilde{Y}_k {-} H_k X|^2
\kern-2pt
\bigg)\!,
\end{equation}
where $\mathcal{X}_m^0$ and $\mathcal{X}_m^1$ denote the subsets of constellation symbols for which the $m$th bit equals 0 or 1, respectively. The presence of $\varepsilon_k$ biases the Euclidean distance metric, leading to overconfident or misleading LLR values.

From a coding perspective, LDPC decoders rely on the consistency and statistical reliability of input LLRs to perform iterative Belief Propagation (BP). Residual NBI introduces structured, non-Gaussian distortions that reduce the mutual information between the transmitted bits and the computed LLRs. Consequently, the decoding process may converge more slowly or stall, resulting in an increased block error rate (BLER) and bit error rate (BER).

This degradation is particularly pronounced when the residual interference is localized to a subset of subcarriers, as the affected LLRs act as unreliable messages within the iterative LDPC decoding graph. Therefore, improving NBI cancellation accuracy directly enhances LLR quality and yields substantial gains in channel decoding performance.

Moreover, residual NBI can lead to the appearance of an error floor at medium-to-high SNRs. In this regime, thermal noise becomes negligible, but structured interference components persist, causing certain bits or subcarriers to experience a consistently degraded effective SNR. As a result, the LDPC decoder receives systematically unreliable LLRs that cannot be corrected through iterative processing, preventing the BER and BLER from decreasing exponentially with SNR.

\subsection{Limitations of Per-Subcarrier Power Estimation}

A theoretical classical solution to avoid these extreme outliers would be to adapt the LLR calculation to the actual noise-plus-interference power at each subcarrier $k$. By effectively replacing the global AWGN variance $\sigma_w^2$ with a per-subcarrier variance $\sigma_{Z,k}^2 = \sigma_w^2 + \mathbb{E}[|\tilde{E}_k|^2]$, the demapper could theoretically ``de-weight" the confidence of subcarriers heavily corrupted by NBI residuals. However, accurately estimating $\sigma_{Z,k}^2$ in real-time is practically and mathematically non-trivial. %
Kalyani \emph{et al.} \cite{kalyaniInterferenceMitigationTurboCoded2008} note that while classical Gaussian-based LLRs are highly vulnerable to non-Gaussian outliers, explicitly modeling the true contaminated Gaussian distribution is practically infeasible due to the complexity of tracking rapidly time-varying interference parameters.

To estimate the residual power $\sigma_{\tilde{E},k}^2 {=} \mathbb{E}[|\tilde{E}_k|^2]$ at subcarrier~$k$, the receiver must isolate $\tilde{E}_k$ from the received signal $Y_k = H_k X_k + \tilde{E}_k + W_k$. Because the transmitted data $X_k$ is a random variable, any real-time estimate of the power at subcarrier $k$ is inherently coupled with the unknown modulation symbol. 

Furthermore, while one could attempt to average power over multiple OFDM symbols to integrate out the data variance, NBI is often transient or non-stationary. More importantly, the residual $\tilde{E}_k$ is a direct function of the estimation error from the mitigation stage (e.g., the fractional frequency offset error $\Delta \alpha$), which varies per symbol based on the behavior of the interferer. 
Thus, $\sigma_{\tilde{E},k}^2$ is not a stationary parameter, but a highly dynamic stochastic variable dependent on the instantaneous estimation error of the NBI canceller. 

Without a precise, \emph{a priori} analytical model of the estimation error distribution produced by the NBI canceller, the receiver cannot differentiate between a naturally noisy subcarrier and one corrupted by structured residual interference. This fundamental intractability motivates the need for a data-driven approach capable of learning the complex statistical footprint of the NBI residuals without requiring explicit variance estimation.

\section{Deep Learning-aided LLR Estimation for Post-mitigation Residual NBI}
\label{sec:llrnet}

As detailed in Section \ref{sec:nbi_estimator}, the \NNEST{} module parameterizes and reconstructs the interference vector $\hat{\mathbf{E}}$ to form the interference-suppressed signal $\tilde{\mathbf{Y}} = \mathbf{Y} - \hat{\mathbf{E}}$. While this subtraction removes the vast majority of the NBI power, the parametric estimation is inherently imperfect. This leaves behind a residual interference component that is highly structured, spectrally correlated,
and profoundly non-Gaussian.

Classical demapping algorithms, such as the widely used max-log approximation, operate under the strict assumption that the post-cancellation noise strictly follows a Gaussian distribution characterized by the background variance $\sigma_w^2$. When a classical Gaussian demapper processes the non-Gaussian residual structure of $\tilde{\mathbf{Y}}$, it misinterprets residual interference peaks as high-confidence data symbols, generating extreme LLR outliers. In a BP decoding graph, these outliers act as mathematical absorbing states, locking the LDPC decoder into false convergence and severely degrading the BER and BLER performance of the system. 

To resolve this challenge, we propose \NNDET{}, a lightweight, deep learning-aided LLR estimation network. Instead of relying on rigid Gaussian assumptions, \NNDET{} utilizes both the cleaned signal $\tilde{\mathbf{Y}}$ and the estimated interference landscape $\hat{\mathbf{E}}$ to dynamically recognize residual patterns and map them into well-calibrated, bounded soft information.

\subsection{LLR-CNet Architecture}
To recover the transmitted bitstream from the mitigated observation, we introduce \NNDET{}. Extending the low-complexity detection methodology originally proposed in \cite{kavvousanosDesignImplementationLowComplexity2025}, \NNDET{} is formulated as an efficient 1D CNN operating directly on the frequency-domain subcarriers. It advances the baseline architecture by integrating the custom circular padding strategy detailed in Section~\ref{subsec:circ-pad}, ensuring robust spatial feature extraction without introducing artificial boundary artifacts at the band edges. While incorporating this structural enhancement to natively handle OFDM spectral wrap-around, \NNDET{} maintains a strict, low-computational footprint, requiring only a single 1D convolutional layer and a pointwise multi-layer perceptron (MLP) head.

The network accepts three distinct inputs: the interference-suppressed subcarrier symbols $\tilde{\mathbf{Y}}\in \mathbb{C}^{N \times 1}$, the reconstructed analytical interference $\hat{\mathbf{E}} \in \mathbb{C}^{N \times 1}$, and the scalar background thermal noise variance $\sigma_w^2$. Providing $\hat{\mathbf{E}}$ alongside $\tilde{\mathbf{Y}}$ is critical; it allows the network to spatially localize areas of heavy cancellation and implicitly scale its detection confidence based on the likelihood of residual estimation errors. To guarantee scale-invariance and gradient stability across arbitrary FFT sizes, the complex input vectors are first power-normalized identically to the \NNEST{} preprocessing stage:
\begin{equation}
    \tilde{\mathbf{Y}}_{\text{in}} = \frac{\tilde{\mathbf{Y}}}{\sqrt{N}}, \quad \hat{\mathbf{E}}_{\text{in}} = \frac{\hat{\mathbf{E}}}{\sqrt{N}}.
\end{equation}

To process the complex-valued arrays using standard real-valued neural network operations, the forward pass is defined through the following sequential stages:
\begin{itemize}
    \item \textbf{Input Formulation:} The normalized complex vectors $\tilde{\mathbf{Y}}_{\text{in}}$ and $\hat{\mathbf{E}}_{\text{in}}$ are decoupled into their real and imaginary components and concatenated along the feature dimension. This forms the initial input feature map $\bm{\psi}^{(0)} \in \mathbb{R}^{N \times 4}$:
    \begin{equation}
        \bm{\psi}^{(0)} = \left[ \Re\{\tilde{\mathbf{Y}}_{\text{in}}\}, \Im\{\tilde{\mathbf{Y}}_{\text{in}}\}, \Re\{\hat{\mathbf{E}}_{\text{in}}\}, \Im\{\hat{\mathbf{E}}_{\text{in}}\} \right].
    \end{equation}
    
    \item \textbf{Feature Extraction:} Because OFDM subcarriers exhibit circular continuity in the frequency domain, standard zero-padding introduces artificial discontinuities. A circular padding layer prepends and appends $P$ adjacent subcarrier features to the sequence using modulo indexing. The padded sequence is then processed by a 1D Convolutional layer employing $F$ filters of kernel size $K$ with a ReLU activation, producing the intermediate spatial map $\bm{\psi}^{(1)} \in \mathbb{R}^{N \times F}$. This operation allows the network to analyze the localized spectral splatter of residual interference across neighboring subcarriers.
    
    \item \textbf{Noise Conditioning:} To allow the network to dynamically scale its confidence based on global channel conditions, the scalar noise variance $\sigma^2_w$ is broadcast across the subcarrier dimension and concatenated to the convolutional feature map, forming the conditioned state tensor $\bm{\psi}_{\text{cond}} \in \mathbb{R}^{N \times (F + 1)}$.
    
    \item \textbf{Soft Demodulation Output:} To directly satisfy the requirements of modern FEC decoders, the network acts as a soft-decision demodulator. The conditioned tensor $\bm{\psi}_{\text{cond}}$ is processed independently across the subcarrier dimension by a hidden Dense layer using a ReLU activation. A final linear Dense layer projects these features into an output tensor $\mathbf{\lambda} \in \mathbb{R}^{N \times n_b}$, where $n_b$ is the number of bits per QAM symbol. Crucially, as mathematically proven in the subsequent training formulation, by omitting a bounding activation (e.g., sigmoid) and allowing the network to output continuous unconstrained logits, these linear outputs naturally converge to the exact LLRs of the coded bits. %
\end{itemize}

\begin{figure}[tb]
    \centering
\includegraphics[width=1\columnwidth]{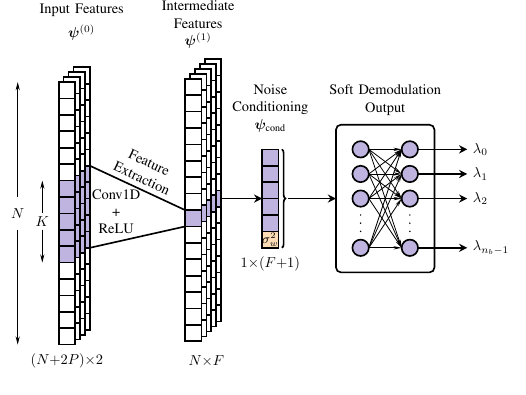}
    \caption{Model Architecture of \NNDET{}. The mathematical stages natively map to the structural blocks, demonstrating the flow from complex decoupling to final LLR generation.}
    \label{fig:llrnet_arch}
\end{figure}

The complete architecture of \NNDET{} is illustrated in Fig.~\ref{fig:llrnet_arch} and its specific structural hyperparameters are outlined in Table~\ref{tab:llrnet_arch}.

\subsection{Training Procedure and Loss Function}
To accurately learn the non-Gaussian statistics of the residual interference, \NNDET{} is trained in a cascaded configuration with the preceding NBI mitigation stage (e.g., OMP-IDS, EOMP-IDS, or \NNEST{}). The mitigation module dynamically processes the noisy dataset to provide the required input features, $\tilde{\mathbf{Y}}$ and $\hat{\mathbf{E}}$. When paired with \NNEST{}, the estimator's weights are strictly frozen to isolate the LLR mapping objective and guarantee training stability. The training hyperparameters mirror Table \ref{tab:training_params}, though \NNDET{} converges more rapidly, requiring only $60,000$ steps.

The \NNDET{} is trained using the binary cross-entropy (BCE) loss function. This objective function is explicitly chosen because it mathematically aligns with the definition of optimal LLRs. The network outputs raw logits $\lambda$, which are passed through a sigmoid activation function during training to yield the predicted probability that the transmitted coded bit $c$ is a `1', defined as
\begin{equation}
    P(c = 1|Y) = \sigma(\lambda) = \frac{1}{1 + e^{-\lambda}}.
\end{equation}
The BCE loss for a single bit is defined as
\begin{equation}
    \mathcal{L}_{\text{BCE}} = -\big[c \log(\sigma(\lambda)) + (1-c) \log(1-\sigma(\lambda))\big].
\end{equation}

It is a well-established property of neural networks that minimizing the expected BCE loss drives the network's output $\sigma(\lambda)$ to converge to the true posterior probability $P(c=1|Y)$. Given this convergence, we can invert the sigmoid function to reveal the physical meaning of the raw logit $\lambda$
\begin{equation}
    e^{-\lambda} = \frac{1 - P(c=1|Y)}{P(c=1|Y)} = \frac{P(c=0|Y)}{P(c=1|Y)}.
\end{equation}
Taking the natural logarithm of both sides yields
\begin{equation}
    \lambda = \log \frac{P(c=1|Y)}{P(c=0|Y)}.
\end{equation}
This formulation proves that by training the network with BCE loss, the linear logits $\lambda$ naturally converge to the exact LLRs. Consequently, the network acts as a structural whitener, learning to suppress overconfident predictions in regions with high residual NBI, thereby providing the LDPC decoder with safe, optimally scaled soft information.

\begin{table}[tb]
    \caption{\NNDET{} Architecture and Parameter Count ($n_b=4$)}
    \label{tab:llrnet_arch}
    \centering
    \includegraphics[width=1\columnwidth]{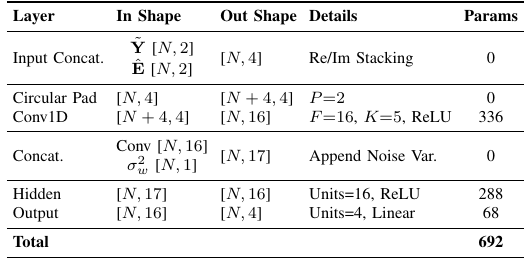}
\end{table}

\begin{figure}[tb]
    \centering
\includegraphics[width=1\columnwidth]{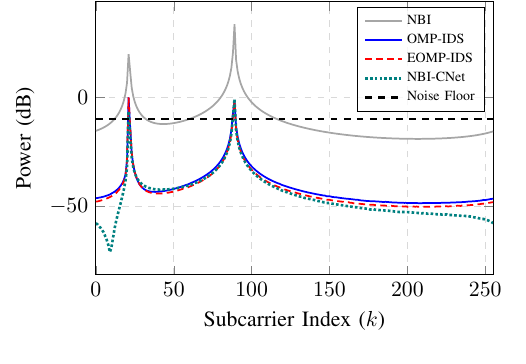}
    \caption{Frequency-domain snapshot of the residual interference power across subcarriers ($\text{Q}{=}2$, $\text{INR}{=}20\ \text{dB}$). Residual spikes remain above the background noise floor, after NBI mitigation.}
    \label{fig:cancellation_snapshot}
\end{figure}

\begin{figure}[tb]
    \centering
\includegraphics[width=1\columnwidth]{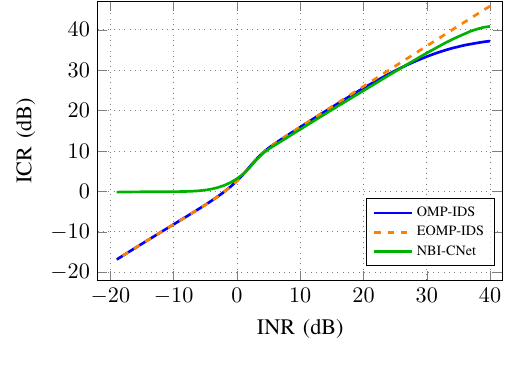}
    \caption{ICR performance of the different algorithms ($\text{Q}{=}8$, $\text{SNR}{=}10\ \text{dB}$, LDPC(1024,512), 16-QAM).}
    \label{fig:cancellation_performance}
\end{figure}

\section{Simulation Results}
\label{sec:simulation_results}

In this section, we evaluate the performance of the proposed \NNEST{} and \NNDET{} architectures. We compare them against classical CS baselines (OMP-IDS and EOMP-IDS) and the Gaussian Max-Log demapper. For simplicity, we assume each OFDM symbol carries exactly one LDPC codeword, meaning the FFT size ($N$) is directly determined by the codeword length. Additionally, we set $d_m{=}4$ to prevent the interference tones from clustering extremely close. Unless otherwise noted, all other simulation parameters follow Table~\ref{tab:training_params}.

\subsection{Interference Cancellation Performance}

To assess the efficacy of the NBI mitigation stage, we adopt the interference cancellation ratio (ICR) metric, $C_{\text{dB}}$, as follows. Let $\mathbf{E}$ denote the true, properly normalized frequency-domain NBI vector for a given OFDM symbol, and let $\hat{\mathbf{E}}$ be the corresponding reconstructed estimate. The per-symbol cancellation depth in decibels (dB) is computed as the ratio of the true interference power to the residual error power and is given by
\begin{equation}
    C_{\text{dB}} = 10 \log_{10} \left( \frac{\|\mathbf{E}\|_2^2 }{\|\mathbf{E} - \hat{\mathbf{E}}\|_2^2} \right).
\end{equation}
To prevent artificial metric deflation from interference-free symbols, the final ICR is averaged exclusively over the valid subset of symbols where the number of active interferers is strictly positive ($Q > 0$).

Fig. \ref{fig:cancellation_snapshot} illustrates a representative frequency-domain snapshot of the residual interference power across subcarriers for a signal corrupted by asynchronous multi-tone NBI ($Q{=}2$, $\text{INR}{=}20$~dB). Before mitigation, the raw interference exhibits massive power peaks that severely exceed the background noise floor, accompanied by extensive spectral leakage across the entire bandwidth. While standard OMP-IDS achieves a baseline level of cancellation, it struggles to perfectly resolve the fractional frequency offsets, leaving a noticeable residual component. The EOMP-IDS algorithm only marginally improves upon this by slightly suppressing the primary interference peaks through its joint iterative searches. In comparison, although \NNEST{} exhibits similar residual spikes above the noise floor, it achieves slightly deeper interference suppression along the spectral tails.

The aggregate cancellation performance evaluated across a wide range of interference-to-noise ratios (INR) is presented in Fig. \ref{fig:cancellation_performance}. A critical architectural advantage of the proposed neural network emerges immediately in the low-INR regime ($\text{INR} {<} 0$ dB), where the interference is submerged beneath the signal and noise power. Despite being provided with perfect prior knowledge of the true number of interferers, the greedy classical algorithms (OMP-IDS and EOMP-IDS) are fundamentally unable to distinguish weak NBI from valid OFDM data. Consequently, they aggressively attempt to cancel the signal of interest, resulting in severe negative cancellation depths that actively destroy the received waveform. In contrast, \NNEST{} leverages its learned sparsity regularizations to intelligently recognize the absence of dominant NBI. It correctly suppresses its parameter predictions, maintaining a safe, flat~$0$~dB baseline that perfectly preserves the underlying signal integrity without inducing artificial distortion.

In the extreme high-INR regime (INR $> 20$ dB), the interference becomes distinctly visible above the noise floor. Here, the computationally exhaustive joint searches of EOMP-IDS allow it to scale linearly to the deepest absolute cancellation floors. While \NNEST{} slightly deviates from this theoretical optimum at the highest INRs, it maintains a highly competitive suppression that strictly outperforms the standard OMP-IDS baseline. More importantly, this slight deviation from the EOMP-IDS optimum is practically insignificant for FEC systems. Once the interference is suppressed sufficiently below the signal power, the downstream \NNDET{} algorithm is fully capable of whitening the remaining residuals.

\subsection{Impact on Soft Information and LLR Calibration}

\begin{figure}[tb]
    \centering
\includegraphics[width=1\columnwidth]{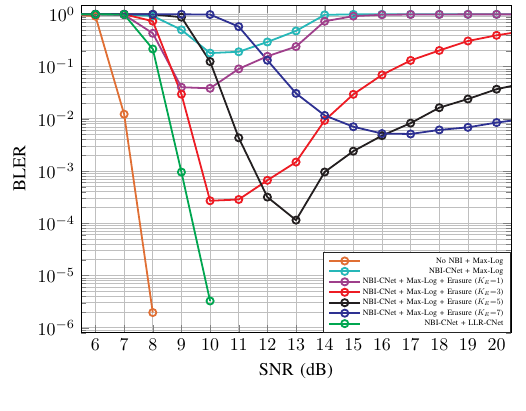}
    \caption{BLER \emph{vs.} SNR performance comparison between LLR-CNet and the LLR erasure method ($\text{Q}{=}24$, $\text{SIR}{=}{-}10\ \text{dB}$, LDPC(2048,1024), 16-QAM).}
    \label{fig:ber-erasure}
\end{figure}

\begin{figure}[tb]
\centering
\includegraphics[width=1\columnwidth]{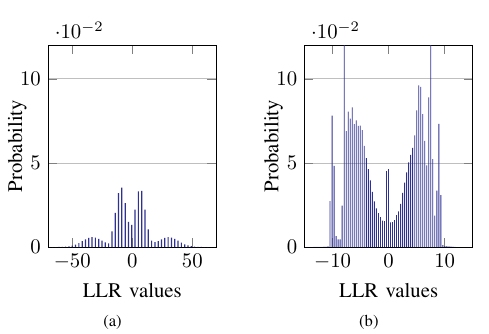}
\caption{LLR values post NBI Mitigation. (a) Gaussian-based Max-log, (b) \NNDET{} ($\text{Q}{=}8$, $\text{SNR}{=}13\ \text{dB}$, $\text{SIR}{=}{-}10$~dB, LDPC(1024,512), 16-QAM).}
\label{fig:llr-distr}
\end{figure}

As established in Section~\ref{sec:llrnet}, the residual interference left behind by the cancellation stage is profoundly non-Gaussian. While optimal classical demappers can theoretically be derived for correlated non-Gaussian noise, doing so requires complex, real-time covariance matrix estimations and inversions that are computationally prohibitive for standard hardware. Consequently, widely deployed demappers operating under strict independent AWGN assumptions exhibit severe performance degradation when processing these structured residuals. 

Fig.~\ref{fig:llr-distr} explicitly quantifies this phenomenon by comparing the probability distributions of the generated LLRs for a scenario with $Q{=}8$ interferers at a SIR of $-10$ dB. When utilizing the Gaussian max-log algorithm (Fig.~\ref{fig:llr-distr}a), the demapper misinterprets the structured residual interference as high-confidence data, generating massive LLR outliers that frequently exceed $\pm 60$. These extreme false-confidence values stall the message-passing convergence in an LDPC belief propagation graph. Conversely, the proposed \NNDET{} (Fig.~\ref{fig:llr-distr}b) seamlessly learns the true statistical distribution of the residuals. It acts as a structural whitener, safely compressing the LLR variance and bounding the soft values within a realistic confidence regime of roughly $\pm 15$. This dynamic calibration prevents decoder lock-up and restores the integrity of the soft bits.

Fig.~\ref{fig:ber-erasure} evaluates the impact of this LLR calibration on the final BLER against a classical frequency-domain blanking strategy, where $K_E$ denotes the erasure window of zeroed subcarriers around each detected interferer. The classical method exhibits a severe trade-off: narrow windows ($K_E {\le} 3$) fail to capture the broader spectral leakage, while wider windows ($K_E {\ge} 5$) indiscriminately puncture too much valid payload data. Both extremes trigger severe high-SNR error floors. In contrast, the joint \NNEST{} and \NNDET{} pipeline structurally whitens the residual interference rather than discarding the affected subcarriers entirely. By actively recovering reliable soft information, the neural approach completely eliminates this erasure penalty, yielding a substantial coding gain of approximately $0.8$~dB at a BLER of $10^{-3}$ over the best-performing classical erasure window, and achieving a steep waterfall curve that converges within $1.7$~dB of the ideal, NBI-free baseline.

\begin{figure*}[ht]
\centering
\includegraphics[width=2\columnwidth]{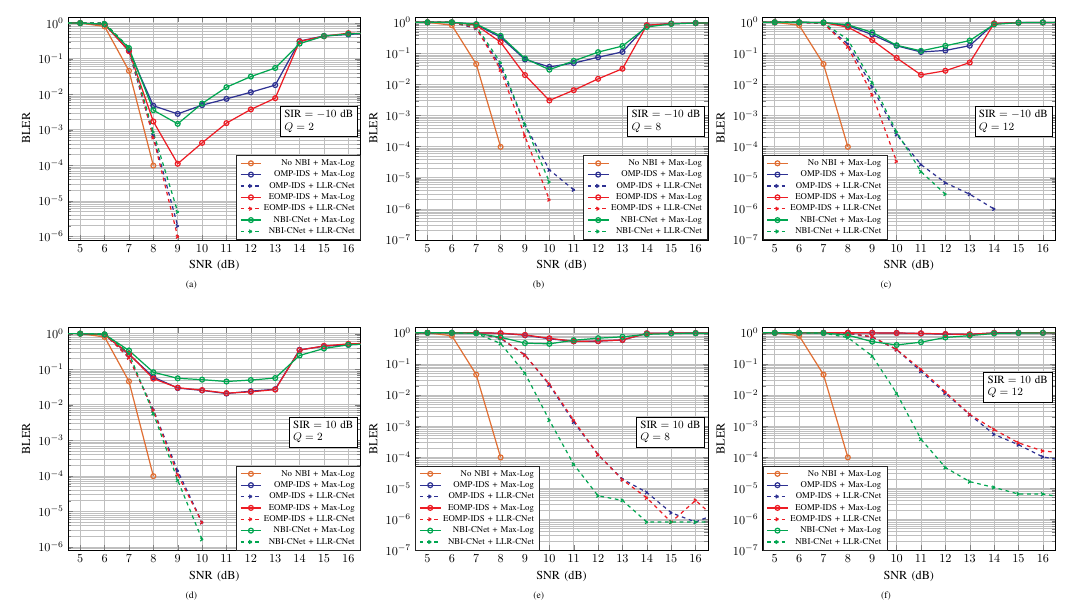}
\caption{BLER \emph{vs.} SNR for the proposed and baseline CS NBI mitigation algorithms under different interference scenarios (LDPC(1024, 512), 16-QAM).}
\label{fig:grid}
\end{figure*}

\begin{figure}[tb]
    \centering
    \includegraphics[width=1\columnwidth]{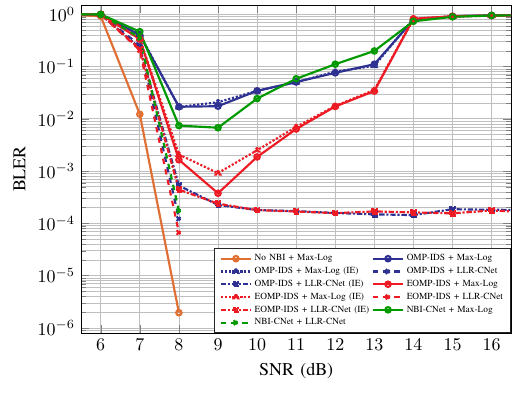}
    \caption{BLER \emph{vs.} SNR performance comparisons ($\text{Q}{=}8$, $\text{SIR}{=}{-}10\ \text{dB}$, LDPC(2048,1024), 16-QAM). The abbreviation IE indicates Imperfect Estimation of the number of active NBI sources.}
    \label{fig:ber2}
\end{figure}

\subsection{End-to-End Channel Decoding Performance}

To evaluate the ultimate impact on the communication link, we analyze the BLER across varying interference severities ($\text{SIR} = 10$ dB and $-10$ dB) and spectral crowding scenarios ($Q=2, 8, 12$), as presented in Fig.~\ref{fig:grid}.

The baseline classical pipelines exhibit notable performance degradation. Combining iterative cancellation (OMP-IDS or EOMP-IDS) with Gaussian Max-Log demapping results in severe error floors. In crowded environments ($Q{=}8$ and~$Q{=}12$), these methods stall at a BLER near $10^{-1}$ even at high SNRs. Such crowded NBI scenarios are prevalent in Broadband PLC systems \cite{martinezmarreroImprovingSoftDecoding2019}.

Replacing the Max-Log demapper with \NNDET{} recovers LDPC decoder performance. As shown across Fig.~\ref{fig:grid}, this substitution eliminates the $10^{-1}$ error floors and restores the exponential BLER decay. Under heavy interference ($\text{SIR}{=}{-}10$~dB), the combined \NNEST{} and \NNDET{} pipeline reaches a target BLER of $10^{-4}$ at SNRs of $8.5$~dB ($Q{=}2$), $9.5$~dB ($Q{=}8$), and $10.5$~dB ($Q{=}12$). Across all tested $Q$ conditions at $\text{SIR}{=}{-}10$~dB, \NNEST{} operates within a $0.2$ to $0.5$~dB SNR margin of the EOMP-IDS + \NNDET{} baseline. This confirms that \NNEST{} achieves comparable interference suppression while bypassing the execution latency of sequential searches.

Under mild interference ($\text{SIR}{=}10$~dB), the limitations of iterative CS methods emerge in dense spectrums. Because the interference power is comparable to the useful signal, the greedy algorithms cannot efficiently distinguish true NBI peaks from valid OFDM data subcarriers. Consequently, the iterative search erroneously targets and subtracts valid signal components, actively destroying the effective payload. While CS methods handle sparse interference ($Q{=}2$), this algorithmic confusion causes severe degradation in crowded grids. At $Q{=}8$ (Fig.~\ref{fig:grid}e), the CS baselines require over $14.5$~dB SNR to reach a $10^{-5}$ BLER. At $Q{=}12$ (Fig.~\ref{fig:grid}f), OMP-IDS and EOMP-IDS exhibit an error floor near $10^{-4}$ at $16$~dB. In contrast, \NNEST{} directly resolves overlapping Dirichlet kernels without confusing them for payload data. It bypasses these error floors, reaching $10^{-5}$ BLER at $11.8$~dB ($Q{=}8$) and $14$~dB ($Q{=}12$). This yields a coding gain of $2$ to over $3$~dB in congested bands, demonstrating vastly improved scalability.

Finally, a major limitation of iterative CS algorithms is their strict dependency on an accurate prior estimation of the interference count ($\widehat{\raisebox{-0.25mm}{\phantom{Q}}}\kern-0.8em Q$). Fig.~\ref{fig:ber2} evaluates an Imperfect Estimation (IE) scenario where the estimator has a minor 0.1\% probability of misestimating $Q$ by $\pm 1$, a condition significantly more optimistic than the practical NBI detection rates reported in literature \cite{huNarrowbandInterferenceCancellation2025}. Despite this favorable setup, this 0.1\% error rate forces both OMP-IDS and EOMP-IDS to exhibit a severe error floor near $2 \times 10^{-4}$, even when paired with \NNDET{}. This degradation is fundamentally algorithmic: underestimating $Q$ leaves a high-power interferer completely unmitigated, whereas overestimating forces the subtraction of non-existent tones, injecting artificial interference into the spectrum. In contrast, \NNEST{} eliminates this vulnerability by natively suppressing interference without requiring an explicit prior estimate $\widehat{Q}$. Consequently, the neural pipeline circumvents the IE error floor entirely, reaching a BLER of $10^{-4}$ at an SNR of $8$~dB and yielding an unbounded relative coding gain in the high-SNR regime.

\subsection{Neural Network Generalization}

A key architectural advantage of the proposed \NNEST{} and \NNDET{} pipelines is their inherent ability to generalize to system configurations not encountered during training, both in terms of spectral dimensions and interference density. 

Although the networks were trained exclusively on a fixed FFT size of $N{=}256$, the structural choice of 1D convolutional layers combined with the power-invariant input normalization allows them to process arbitrary OFDM symbol lengths. To achieve this dimension-agnostic generalization, a deterministic scaling factor must be applied to the predicted interference gain. As explicitly defined in the closed-form interference model in \eqref{eq:nbi_freq_closed_form}, the frequency-domain amplitude of the NBI scales proportionally with $1/\sqrt{N}$. Because the network's raw amplitude prediction $\hat{g}$ is implicitly calibrated to the spectral energy density of the training FFT size $N$, performing inference on a novel grid size $N'$ requires compensating for this fundamental analytical dependence. To preserve the correct physical interference power during time-domain reconstruction, the predicted gain must be scaled as
\begin{equation}
\hat{g}' = \hat{g} \sqrt{\frac{N'}{N}}.
\end{equation}

By applying this constant scalar, the neural pipeline successfully generalizes to an extended LDPC(2048, 1024) codeword mapped over $N' {=} 512$ subcarriers, as demonstrated in Fig.~\ref{fig:ber2}. Despite operating on double the spectral bandwidth relative to the training dataset, the integrated \NNEST{} and \NNDET{} architecture maintains robust decoding performance. In contrast, the classical max-log approaches continue to exhibit an uncorrectable BLER of $1.0$. This robust, length-invariant scaling confirms that the network has fundamentally learned the localized physical properties of the continuous fractional frequency offsets and the underlying spectral leakage, as governed by $\mathcal{D}_N$ in \eqref{eq:nbi_freq_closed_form}, rather than merely overfitting to a specific discrete subcarrier grid.

Beyond spectral dimensions, the neural pipeline demonstrates exceptional generalization regarding interference density. While the models were trained exclusively on environments containing up to a maximum of $8$ interferers ($Q {\in} [0, 8]$), their fully convolutional nature processes the entire frequency spectrum simultaneously. This wide receptive field allows \NNEST{} to independently identify and isolate multiple NBI sources without relying on a predefined number of sequential loops. Consequently, when evaluated in denser spectral crowding scenarios (e.g., $Q{=}12$, as presented in Fig.~\ref{fig:grid}), \NNEST{} successfully suppresses the elevated interference, without requiring retraining. Likewise, \NNDET{} is able to generalize to arbitrary input FFT size and NBI density, continuing to produce well-calibrated soft information.

\section{Computational Complexity}
\label{sec:complexity}

To benchmark the computational burden of the baseline classical algorithms against the proposed neural network, the complex-valued asymptotic counts provided in existing literature \cite{huNarrowbandInterferenceCancellation2025} must be translated into real floating-point operations (FLOPs). In standard digital signal processing hardware architectures, a single complex addition requires 2 real FLOPs, while a single complex multiplication requires 6 real FLOPs (4 real multiplications and 2 real additions).

\begin{table}[tb]
\centering
\caption{Real FLOPs Comparison of Baseline NBI Cancellation Algorithms vs. \NNEST{}}
\label{tab:flops_comparison}
\includegraphics[width=1\columnwidth]{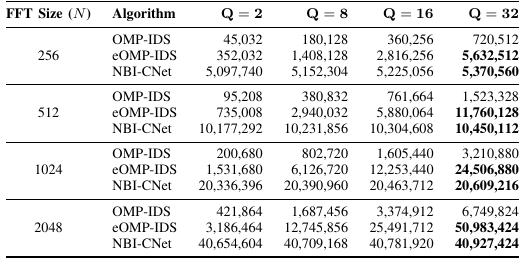}
\end{table}

\begin{table}[tb]
\centering
\caption{FLOP Count of the \NNDET{}}
\label{tab:flops_nndet}
\includegraphics[width=0.4\columnwidth]{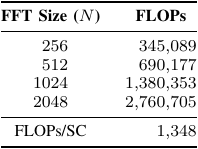}
\end{table}

\subsection{Complexity of OMP-IDS and EOMP-IDS}
Based on the operations detailed in Table IV of \cite{huNarrowbandInterferenceCancellation2025}, the total real FLOPs for the OMP-IDS and EOMP-IDS algorithms per OFDM symbol are derived using the number of active narrowband interferers ($Q$) and the FFT size ($N$). Let $K_1$ denote the number of standard dichotomous search iterations, $K_2$ the joint search iterations, $P$ the number of adjacent evaluation points, and $\bar{P}$ the average number of joint-search iterations triggered across the spectrum. The parametric computational complexities can be expressed as
\begin{align}
    \mathcal{C}_{\text{OMP-IDS}} & = Q \Big[ 6 \big(2K_1 N + \frac{N}{2}\log_2 N \big) + \notag \\
    &\qquad + 2 \big( 2K_1(N-1) + N\log_2 N \big) \Big] \\
\mathcal{C}_{\text{EOMP-IDS}} & =  Q \Big[ 6 \big( 2N(K_1 + P K_2) + \notag \\
&\qquad (\bar{P}+1)\frac{N}{2}\log_2 N \big) \nonumber \\ &+ 2 \big( 2(N-1)(K_1 + P K_2) \notag \\
&\qquad\qquad + (\bar{P}+1)N\log_2 N \big) \Big].
\end{align}
By grouping like terms  these expressions reduce to
\begin{align}
    \mathcal{C}_{\text{OMP-IDS}} & = Q \Big[ 5N\log_2 N + 16K_1 N - 4K_1 \Big] \\
    \mathcal{C}_{\text{EOMP-IDS}} & = Q \Big[ 5N(\bar{P}+1)\log_2 N + 16N(K_1 + P K_2) \notag \\&\qquad\qquad- 4(K_1 + P K_2) \Big].
\end{align}

Following the hyperparameters evaluated in \cite{huNarrowbandInterferenceCancellation2025} for optimal cancellation depth ($K_1{=}3$, $K_2{=}5$, $P{=}5$, and an average trigger rate $\bar{P}{=}5$), these expressions simplify to exact polynomial FLOP counts per NBI source
\begin{align}
    \mathcal{C}_{\text{OMP-IDS}} & = Q \big( 5N \log_2 N + 48N - 12 \big) \label{eq:flops_omp} \\
    \mathcal{C}_{\text{EOMP-IDS}} & = Q \big( 30N \log_2 N + 448N - 112 \big). \label{eq:flops_eomp}
\end{align}

\subsection{Complexity of NBI-CNet}

To derive the comparable FLOPs for the proposed \NNEST{} architecture, we adopt the standard deep learning convention where one Multiply-Accumulate (MAC) operation contributes two FLOPs. 

Based on the network architecture established in Section~\ref{sec:nbi_estimator}, the total parameter count is distributed across the feature extraction backbone ($P_{\text{bb}} {=} 7,680$), the gain head ($P_{g} {=} 2,241$), the fractional frequency head ($P_{\alpha} {=} 2,241$), and the phase head ($P_{\theta} {=} 2,306$). In a standard, unconditional execution model, the network evaluates all feature maps and outputs across the entire subcarrier grid ($N$). The total complexity is therefore
\begin{align}
    \mathcal{C}_{\text{\NNEST{}-Dense}} &= 2 N (P_{\text{bb}} + P_{g} + P_{\alpha} + P_{\theta}) \notag \\
    &= 2 N (14,468)  = 28,936 N.
\end{align}
Unlike the CS algorithms in \eqref{eq:flops_omp} and \eqref{eq:flops_eomp}, the execution of \NNEST{} is completely independent of $Q$. While this guarantees a fixed FLOP budget independent of interference density, it inherently forces the network to execute redundant computations on interference-free subcarriers.

\subsection{Sparsity-driven Reduced Complexity}

To optimize execution efficiency without sacrificing the highly parallelized, deterministic latency of a feed-forward architecture, \NNEST{} can employ dynamic conditional routing. Because multi-tone NBI is fundamentally sparse in the frequency domain ($Q {\ll} N$), the detailed extraction of the fractional offset ($\hat{\alpha}$) and phase ($\hat{\theta}$) can be computationally bypassed for the vast majority of the spectrum. 

Under this sparse execution paradigm, only the backbone and the primary gain head must be evaluated across all $N$ subcarriers to detect the locations of the interferers. This forms the static computational baseline of the network
\begin{align}
    \mathcal{C}_{\text{static}} &= 2 N (P_{\text{bb}} + P_{g}) \notag \\
    &= 2 N (9,921) = 19,842 N.
\end{align}
Once the interference profile is established, the frequency and phase heads are triggered exclusively at the indices where active NBI is detected ($\hat{g} {>} 0$). Assuming the network correctly identifies the $Q$ active interferers, the conditional operational cost becomes strictly proportional to the physical interference density
\begin{align}
    \mathcal{C}_{\text{dynamic}} &= 2 Q (P_{\alpha} + P_{\theta}) \notag \\
    &= 2 Q (4,547) = 9,094 Q.
\end{align}
The final, optimized FLOP requirement for a single forward pass of \NNEST{} is the sum of the static and dynamic phases
\begin{equation}
    \mathcal{C}_{\text{\NNEST{}-Sparse}} = 19,842 N + 9,094 Q.
\end{equation}
By leveraging the structural sparsity of the gain mask, the conditional architecture dynamically eliminates $9,094 (N - Q)$ redundant floating-point operations per inference. %

\subsection{Complexity of LLR-CNet}

Following the same MAC conversion methodology, we establish the computational complexity of the neural soft demapper. While \NNEST{} utilizes dynamic routing to optimize its comparatively heavier feature extraction, \NNDET{} requires no such architectural sparsity due to its exceptionally shallow design. 

As detailed in Table~\ref{tab:llrnet_arch}, the entire estimator is strictly constrained to a parameter count of just $P_{\text{\NNDET{}}}{=}692$, distributed across a single 1D convolutional layer and two pointwise linear transformations. Because reliable soft information must be generated for the entire LDPC codeword to prevent downstream decoder failure, \NNDET{} executes unconditionally across the entire subcarrier grid ($N$). Consequently, its execution scales strictly linearly with the FFT size.

Based on the operations defined in the architecture, the estimator requires an average of $1,348$ FLOPs per subcarrier (as detailed in Table~\ref{tab:flops_nndet}). The total complexity for a single forward pass is thus given by
\begin{align}
    \mathcal{C}_{\text{\NNDET{}}} &\approx 1,348 N.
\end{align}

Because this complexity is entirely independent of the interference density $Q$, it maintains the strict, predictable latency profile of the feed-forward pipeline. %

\subsection{Complexity Comparison}

\begin{figure}[tb]
    \centering
    \includegraphics[width=1\columnwidth]{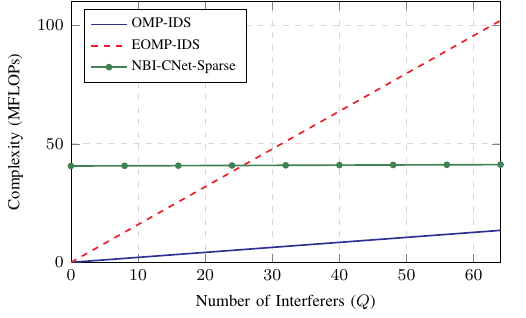}
    \caption{Computational complexity (FLOPs) scaling versus the number of active interferers ($Q$) for $N=2048$. The conditional routing of \NNEST{} provides a nearly flat scaling footprint, bypassing the steep computational scaling penalty of EOMP-IDS in congested bands ($Q > 26$).}
    \label{fig:flops_vs_q}
\end{figure}

Table~\ref{tab:flops_comparison} details the theoretical FLOP formulas across varying FFT sizes ($N$) and sparsity levels ($Q$). To clearly illustrate the practical impact of these scaling parameters, Fig.~\ref{fig:flops_vs_q} visualizes the computational footprint of the algorithms across an extended sparsity range up to $Q{=}64$ for a large grid size ($N{=}2048$). 

While standard OMP-IDS maintains a relatively low computational footprint, the joint dichotomous searches of EOMP-IDS introduce a severe linear scaling penalty. As visualized by the steep slope in Fig.~\ref{fig:flops_vs_q}, EOMP-IDS demands over 101.9 million FLOPs as the spectral environment becomes heavily congested ($Q{=}64$). Crucially, this computational load consists entirely of variable-latency sequential iterations, posing a significant bottleneck for strict real-time baseband processing. Furthermore, the OMP-IDS and EOMP-IDS FLOP counts omit the overhead of the mandatory model-order ($\widehat{Q}$) selection stage. Prior art employs a separate neural network (NDNet) requiring $6.144 \times 10^5$ FLOPs, which yields roughly 98\% estimation precision under high INR conditions \cite{huNarrowbandInterferenceCancellation2025}. Because eliminating the resulting baseline error floors entirely demands near-perfect estimation accuracy across all operational regions, a significantly more complex and computationally heavy $\widehat{Q}$-detection stage would be required in practice.

Conversely, \NNEST{} establishes a static baseline due to its full-grid feature extraction, which is visually evident at $Q \approx 0$. However, it achieves highly efficient scaling via dynamic conditional routing. As $Q$ increases, the network's dynamic FLOP overhead grows almost imperceptibly, producing a nearly flat scaling trajectory. As demonstrated by the intersection point in Fig.~\ref{fig:flops_vs_q}, \NNEST{} becomes strictly more efficient than EOMP-IDS once the spectrum exceeds $Q{=}26$ active interferers. At the maximum evaluated density of $Q{=}64$, \NNEST{} requires roughly 60\% fewer absolute operations than EOMP-IDS (41.2M \emph{vs.}\ 101.9M FLOPs). Furthermore, because \NNEST{} operates entirely without prior knowledge of $Q$, it completely bypasses the complexity and accuracy limitations of a separate estimation stage, while evaluating all active subcarriers simultaneously to guarantee a predictable, parallelized execution profile.

Moreover, evaluating computational complexity solely through raw FLOPs fails to capture the practical deployment efficiency of the neural architecture. CS iterative algorithms require high-precision floating-point arithmetic to ensure numerical stability during matrix pseudo-inversions. In contrast, deep learning models are exceptionally resilient to low-precision arithmetic and inherently exploit highly efficient, domain-specific hardware accelerators. State-of-the-art deployment optimizations, such as post-training quantization (PTQ) to FP16 or INT8, seamlessly compress the computational footprint of \NNEST{}. By leveraging specialized parallel units (such as mixed-precision Tensor Cores) the neural estimator achieves massive reductions in effective latency and memory bandwidth, significantly widening its real-world advantage over classical algorithms.

\begin{table}[t]
\caption{TensorRT Average Execution Time for \NNEST{} and \NNDET{}}
\label{tab:trt_precision_benchmarks}
\centering
\includegraphics[width=1\columnwidth]{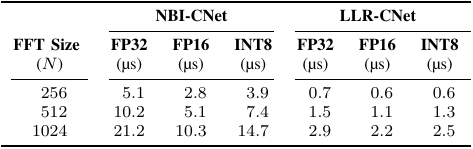}
\end{table}

\subsection{Execution Time}
\label{subsec:exectime}

Table \ref{tab:trt_precision_benchmarks} details the execution latency per OFDM symbol for both the \NNEST{} and \NNDET{} architectures, evaluated over 1,000 iterations (batch size 4096) on an Nvidia DGX Spark using TensorRT~\cite{zhouTensorRTImplementationsModel2023}. Across all precision profiles, execution latency scales linearly with $N$ for both networks. Transitioning the heavier \NNEST{} backbone from FP32 to FP16 yields a near-optimal $2\times$ latency reduction (e.g., from 21.2~\textmu{}s to 10.3~\textmu{}s at $N{=}1024$), successfully halving VRAM bandwidth requirements while efficiently saturating the Grace Blackwell Tensor Cores. Similarly, the highly lightweight \NNDET{} architecture exhibits an exceptionally low computational footprint, executing in under 3~\textmu{}s across all configurations, with FP16 providing a proportional latency reduction (decreasing from 2.9~\textmu{}s to 2.2~\textmu{}s at $N{=}1024$). Conversely, applying INT8 PTQ introduces a noticeable performance regression in both models. Because these lightweight 1D convolutional architectures are inherently memory-bound, the computational overhead of the inserted Quantize/Dequantize (Q/DQ) nodes (combined with the data reformatting overhead required to align tensors for 8-bit execution) outweighs the raw speedup of 8-bit arithmetic. Consequently, INT8 latencies consistently fall between the FP16 and FP32 baselines, establishing the pure FP16 TensorRT engine as the optimal deployment configuration for the end-to-end joint pipeline.

\section{Conclusion}
\label{sec:conclusion}
In this paper, we proposed a joint DL pipeline for the mitigation and soft demodulation of narrowband interference in OFDM systems. We introduced \NNEST{}, a physics-informed convolutional neural network that estimates multi-tone interference parameters in a single, parallelizable forward pass by explicitly modeling Dirichlet kernel spectral leakage. Operating natively without prior knowledge of the active interferer count, its computational footprint scales additively rather than multiplicatively. This enables it to fall strictly below the complexity of EOMP-IDS in high-density scenarios--achieving up to a $60\%$ reduction in absolute operations for large grid sizes ($N{=}2048, Q{=}64$)--while completely eliminating the sequential latency bottlenecks of iterative CS methods.

To address the non-Gaussian residual interference that degrades downstream LDPC decoders, we developed \NNDET{}, a structural whitener that dynamically recalibrates soft information and eliminates the error floors produced by classical Gaussian-based demappers.

Simulation results confirmed that the joint \NNEST{} and \NNDET{} architecture tracks closely to the optimal baseline under severe interference ($\text{SIR}{=}{-}10$\,dB). Crucially, under mild interference conditions ($\text{SIR}{=}10$\,dB), the framework provides significant performance gains by avoiding the signal-peak confusion that causes conventional greedy algorithms to erroneously target and destroy valid payload data. Furthermore, the architecture entirely circumvents the error floors typically triggered in classical pipelines by minor model-order estimation errors, and generalizes to arbitrary FFT grids without retraining via a closed-form gain rescaling. Hardware benchmarks further confirmed real-time viability, with FP16 TensorRT inference offering the optimal latency-performance trade-off for next-generation physical-layer deployments.

\balance{}
\bibliographystyle{IEEEtran}
\bibliography{bibliography}

\end{document}